# MARQ: A Modular, LLM-Agnostic Requirement Quality Assessor Infused with Systems Engineering Knowledge

Jannatul Shefa[1], Alejandro Salado[2], Paul Wach[2], and Taylan G. Topcu[1]

[1]Grado Department of Industrial and Systems Engineering, Virginia Tech, Blacksburg, Virginia, USA

[2]Department of Systems and Industrial Engineering, University of Arizona, Tucson, Arizona, USA

**Correspondence:** Jannatul Shefa (shefa@vt.edu)

## Abstract

Requirements engineering (RE) is foundational to systems engineering (SE), as requirement defects can propagate downstream and adversely affect development outcomes. Although large language models (LLMs) are promising for requirement quality assessment, prior work shows that off-the-shelf models miss a large share of expert-identified defects, perform unevenly across quality criteria, and vary across executions. This study investigates whether infusing expert-derived SE knowledge can improve LLM-based requirement quality evaluation beyond off-the-shelf capability. We present MARQ, a modular, LLM-agnostic prototype that supplements LLMs with a criterion-specific knowledge stack operationalizing INCOSE requirement quality criteria and evaluated against off-the-shelf baselines on an expert-labeled NASA requirement set using Claude Opus 5, GPT-5.6 Sol, and the open-weight Gemma 3 12B, with 100 executions per model. MARQ achieves median F1 scores of 0.80 for Claude Opus 5, 0.62 for GPT-5.6 Sol, and 0.53 for Gemma 3 12B, compared with off-the-shelf baselines of 0.55, 0.52, and 0.18, respectively. However, the nature of these gains is model-dependent. Claude Opus 5 improves defect detection and false-alarm control; GPT-5.6 Sol gains detection and better execution-level consistency at the cost of more false alarms; and Gemma 3 12B reaches the performance of off-the-shelf proprietary models, although it remains highly variable. Criterion-level analysis shows gains concentrated in Correctness and Conformance, trade-offs between misses and false alarms within individual criteria, and model-dependent responses across several criteria. These findings position SE knowledge infusion as a promising model-agnostic pathway for strengthening trustworthy LLM-based RE support while underscoring the need for criterion-specific interventions and human-in-the-loop review.

## 1. Introduction and Background

Requirements are foundational to systems engineering (SE) because they identify the goals to be achieved by the envisioned system (Van Lamsweerde, 2000) and translate them into well-defined statements that constrain and guide subsequent system development (Nuseibeh & Easterbrook, 2000; Simon, 1973). As complex system development progresses through tightly coupled and interdependent activities (Simon, 1995; Hazelrigg, 1998), errors introduced during requirements engineering (RE) can propagate into later lifecycle phases (Mahaju et al., 2023) and become increasingly costly to correct, contributing to design rework, schedule delays, and cost overruns (Boehm & Papaccio, 1988; Peña & Valerdi, 2015; Sikdar & Payyazhi, 2014). Relative to correcting an error during the requirements phase, correction costs can increase to approximately 8× during design, 20× during manufacturing, 100× during integration and testing, and more than 1,500× once the system enters operation (Stecklein et al., 2004; Boehm et al., 2008). Consequently, assuring requirement quality is a critical SE activity, but one that remains time-consuming and dependent on expert engineering judgment (Courbis et al., 2026).

The rapid advancement of artificial intelligence (AI), particularly large language models (LLMs), has created new opportunities to support RE activities (Cheng et al., 2026). This growing interest builds on a longer history of applying AI for RE (AI4RE) partly because requirements are typically expressed in natural language (Kassab et al., 2014). In addition to that, the software engineering community, where RE has long been a critical concern (Cheng et al., 2026; Vasudevan & Reddivari, 2025), played a major role in advancing natural language processing (NLP) methods for RE (Dalpiaz et al., 2018). Early NLP methods lacked contextual awareness (Hayes-Roth, 1985); advances in learning-based approaches (Baker, 1975; Jelinek et al., 1977), deep learning (Elman, 1990), and ultimately transformer architectures progressively improved contextual language processing and sequence prediction, resulting in the substantial performance gains associated with modern LLMs (Ashish, 2017; Hochreiter & Schmidhuber, 1997; Sutskever et al., 2014). Recent reviews (Norheim et al., 2024; Sonbol et al., 2022; Zhao et al., 2022) organize AI4RE research into four categories: requirements generation, analysis, translation, and quality assessment. Quality assessment, which examines whether requirements satisfy established quality criteria, such as Necessity, unambiguity, conformance, and feasibility (Hull et al., 2010; INCOSE, 2023); is the core focus of this study.

Research on AI-based requirement quality assessment spans both individual requirements and requirements sets. At the individual level, studies have examined ambiguity, completeness, writing-guideline adherence, and syntactic quality incorporating AI support. For example, natural language inference has been used to detect specification defects with an F1 score above 80%, outperforming NoRBERT, ChatGPT, probabilistic term-indicator, and prompt-based classification (Fazelnia et al., 2024). Luitel et al. (2023) used unsupervised learning mechanisms to assess completeness (Luitel et al., 2023). Jafari et al. (2023) showed that specialized encoder-based models perform better than GPT-3.5 and GPT-4 models for ambiguity resolution (Jafari et al., 2023). Hybrid pipelines combining SpanBERT, named entity recognition, rules, and Random Forest classifiers have targeted several ambiguity types (Izhar et al., 2025; Talha et al., 2025). More recently, Lubos et al. (2024) showed that Llama 2 has the ability to identify quality issues, explain them, and suggest revisions against ISO 29148 criteria (Lubos et al., 2024).

At the requirement-set level, research has focused mostly on incompleteness, redundancy, conflict, and inconsistency using AI-based approaches. For example, Malik et al. (2025) used semantic similarity to detect redundancy and conflict in requirements (Malik et al., 2025), while GPT-4 achieved a precision range of 41% to 61% in detecting conflicts, ambiguities, and inconsistencies among requirements (Mahbub et al., 2024). Fantechi et al. (2023) similarly found that ChatGPT could assist with inconsistency detection but did not outperform expert review (Fantechi et al., 2023). Finally, it was documented that LLMs combined with clustering and semantic similarity can support quality improvement in large SE projects, particularly for understandability, ambiguity, and redundancy analysis (Melo, 2026).

To better characterize the reliability of LLM-based requirement quality assessment, our prior research systematically benchmarked ten off-the-shelf LLMs spanning the OpenAI and Anthropic model families and five model generations (Shefa et al., 2026). That benchmark found substantial missed detections and false alarms, particularly weak performance for criteria requiring contextual and systems-level judgment (Correct, Necessary, Complete), persistent execution-to-execution variability, non-monotonic improvement across model generations, and insensitivity to sampling temperature (where applicable). It concluded that general-purpose LLM capability should not be equated with dependable engineering judgment (see also Heesch et al., 2025; Zhou et al., 2026), and that future gains are more likely to come from targeted interventions, such as domain-specific knowledge integration, than from waiting for newer models. The present study is the direct test of that recommendation.

One emerging direction is to supply LLMs with project- and domain-specific documentation rather than relying on pretrained model knowledge (Arora et al., 2024). Poulsen et al. (2026) improved ambiguity detection and reduced false positives by supplying project information extracted from NASA James Webb Space Telescope documentation (Poulsen et al., 2026). Similarly, Aysan et al. (2026) used retrieval-augmented generation over domain documentation to detect ambiguity and conflict smells across several LLMs, with gains for ambiguity but mixed effects for conflicts (Aysan et al., 2026). Jin et al. (2025) proposed a multi-agent requirements development framework in which agents are equipped with domain and RE expert knowledge extracted from professional sources, with humans retained for review (Jin et al., 2025). These studies share an important feature: they augment the model with knowledge about *what the system is*. We contend that far less is known about the effect of augmenting the model with knowledge about *how quality criteria should be judged by systems engineers*.

These questions matter for large-scale complex engineered systems in particular, where requirement evaluation depends on interdisciplinary SE knowledge, physical and operational constraints, and broader system context rather than linguistic cues alone (Hull et al., 2010; Vincenti, 1990). LLMs cannot readily substitute for the intuition, experience, and specialized knowledge of domain experts (Arora et al., 2024); they exhibit uneven domain knowledge, hallucinations, limitations in technical reasoning, and stochastic output behavior across repeated executions (Kaddour et al., 2023; Shefa et al., 2026; Topcu et al., 2025; Zhang et al., 2025). These characteristics manifest as both false alarms and missed defects (i.e., Type I and II errors), so aggregate performance alone is an insufficient basis for judging their suitability (Cui & Alexander, 2026; Norheim et al., 2024). These concerns become even more consequential as agentic AI approaches gain traction in RE and SE, because LLM-based evaluators may increasingly function as specialized components within larger multi-agent workflows (Hu et al., 2025; Sapkota et al., 2025).

To address these gaps, this study presents and evaluates Modular Assessor of Requirement Quality (MARQ), a prototype that infuses expert-derived, criterion-specific SE knowledge into general-purpose LLMs. MARQ is designed to be LLM-agnostic. Its knowledge stack, prompt assembly, and output parsing are decoupled from the model behind a single adapter, so that the same SE knowledge can be attached to a proprietary frontier model or to an open-weight model hosted locally. We evaluate MARQ against off-the-shelf baselines using three heterogeneous LLMs, Claude Opus 5, GPT-5.6 Sol, and Gemma 3 12B, with 100 independent executions per model under each condition. Specifically, we ask:

**RQ1:** How and to what extent does MARQ affect requirement quality evaluation performance relative to an off-the-shelf LLM baseline, and does the effect hold across model families?

**RQ2:** How do the effects of MARQ vary across INCOSE requirement quality criteria?

The study makes four contributions: (i) a reusable, criterion-specific SE knowledge stack that operationalizes the INCOSE individual-requirement quality criteria into testable rules; (ii) a modular, LLM-agnostic prototype architecture that applies that knowledge without fine-tuning, retrieval, or model-specific prompt engineering; (iii) empirical evidence on how knowledge infusion changes overall performance, misses, false alarms, and execution-level consistency across model families; and (iv) criterion-level evidence on where knowledge infusion helps, where it trades one error type for another, and where it does not help.

The remainder of this paper is organized as follows. Section 2 describes the research methodology, and Section 3 presents the results for the overall comparison between the off-the-shelf and SE knowledge-infused conditions, followed by criterion-level findings. Section 4 synthesizes the findings, including performance tradeoffs, model-specific behavior, and considerations for trustworthy AI-assisted RE. Section 5 concludes the paper by summarizing the study's contributions, limitations, and directions for future research.

## 2. Methodology

This study follows a six-step methodology, illustrated in Figure 1. First, a requirement set was adapted from a SE problem. Second, an expert-derived ground truth was established for the selected INCOSE requirement quality criteria, while a criterion-specific SE knowledge stack was developed in parallel. Third, two experimental conditions were defined: an off-the-shelf LLM baseline and the MARQ prototype. Fourth, automated n8n workflows executed both conditions repeatedly for each model and stored the evaluations. Finally, the LLM-generated evaluations were compared against ground truth. Steps 1, 2, and 6 follow Shefa et al. (2026) and are summarized here for completeness; Sections 2.3, 2.4, and 2.5 describe Steps 3, 4, and 5 in detail that are the point of departure.

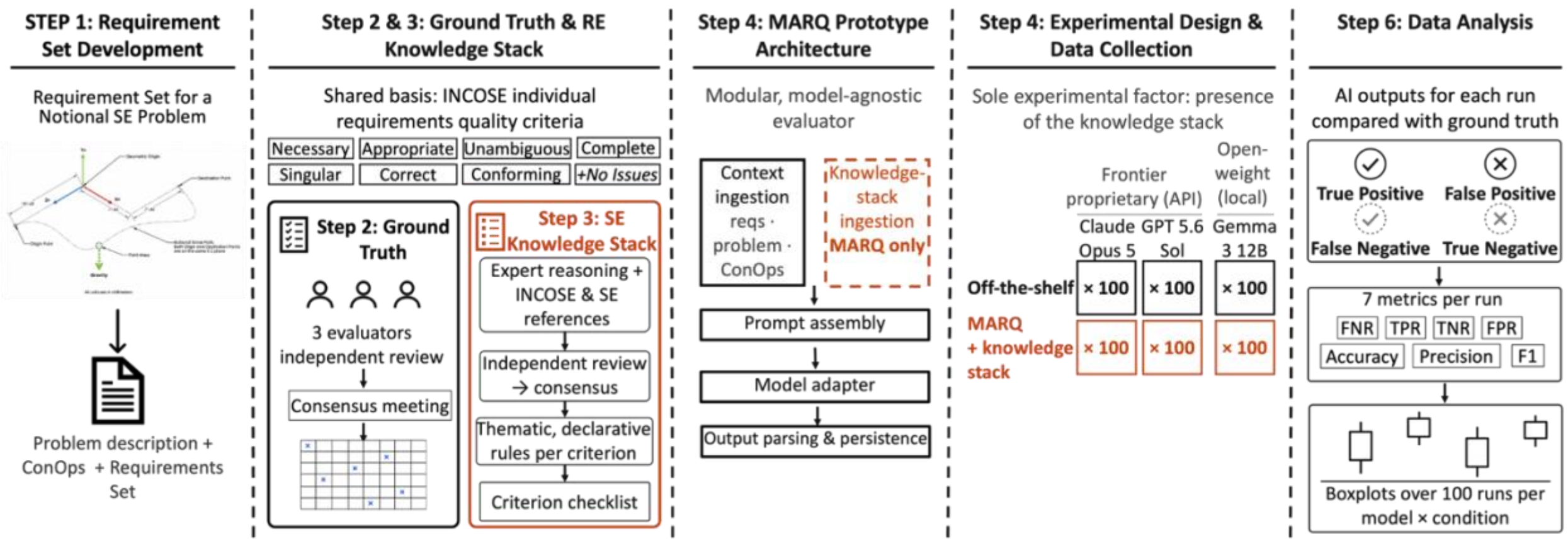

**Figure 1.** Overview of the research methodology.

### 2.1 Step 1: Requirement Set Development

The requirement set used in this study was adapted from the Simple Deployment Mechanism (SDM) challenge of the National Aeronautics and Space Administration (NASA)'s Astrobee Challenge Series (Szajnfarber et al., 2020, 2023). The system of interest is a robotic point-mass manipulator intended to operate on the International Space Station that must move a point mass between specified locations and return it, as defined by a problem description and a Concept of Operations (ConOps). Thirteen requirements were randomly selected from the original SDM dataset for ground truth generation. This set was chosen for its well-defined system context and relatively balanced quality, and because independent developers produced working solutions from it that were subsequently verified by NASA experts, which gives confidence that it is a realistic basis for studying requirement quality evaluation. The requirement set was further categorized using Salado's work on Max-Neef's existential categories (Salado & Nilchiani, 2014), yielding four groups: functional, performance, environmental, and resource requirements.

### 2.2 Step 2: Ground Truth and Knowledge Base Generation

Each requirement was assessed against the INCOSE individual-requirement quality criteria listed in Table 1 (INCOSE, 2023). Feasibility and Verifiability were excluded because the available problem context does not describe organizational capabilities well enough to judge them with confidence, and a No Issues label was used for requirements with no identified defect. The analysis is limited to individual-requirement quality as a first step toward an AI-supported evaluation capability.

The ground truth was developed through evaluations conducted by a panel of three members of the research team with differing levels of RE expertise, including two seasoned reviewers with decades of experience. Each independent evaluator reviewed every requirement against the selected quality criteria and assigned a binary judgment to each requirement–criterion combination. To reduce potential bias and limit intercoder variability (Miles, 1994), this initial phase of the assessments was conducted separately. The evaluators then met to compare their judgments, reconcile differences in how the criteria were interpreted, and resolve disagreements through discussion. The final ground truth used in the study was established through this consensus process.

**Table 1.** Quality Evaluation Criteria for Individual Requirements from INCOSE Handbook

| Criterion | Definition |
|---|---|
| Necessary | The requirement statement defines capability, characteristic, constraint, or quality factor needed or required to satisfy a lifecycle concept, need, source, or higher-level requirement. |
| Appropriate | The specific intent and amount of detail of the need or requirement statement is appropriate to the level (the level of abstraction, organization, or system architecture) of the entity to which it refers. |
| Unambiguous | Need and requirement statements must be stated such that their intent is clear and can be interpreted in only one way by all intended audiences. |
| Complete | The need statement sufficiently describes the necessary capability, characteristic, constraint, conditions, or quality factor to meet the lifecycle concept or source from which it was transformed. The requirement statement sufficiently describes the necessary capability, characteristic, constraint, conditions, or quality factor to meet the need, source, or higher-level requirement from which it was transformed. |
| Singular | The need or requirement statement should state a single capability, characteristic, constraint, or quality factor. |
| Correct | The need statement must be an accurate representation of the lifecycle concept or source from which it was transformed. The requirement statement must be an accurate representation of the need, source, or higher-level requirement from which it was transformed. |
| Conforming | Statements and expressions of individual needs and requirements should conform to an approved standard pattern and style guide or standard for writing and managing needs and requirements. |

This consensus process produced a unified ground truth label for every requirement–criterion pair in the evaluation matrix. Each cell was coded as either positive, indicating that a quality issue was present (marked as **'**x**')**, or negative, indicating that no issue was identified (left empty). This resulted in 104 binary judgments across 13 requirements and eight criteria. This ground truth was then used as the basis for evaluating and comparing the AI-generated assessments across the experimental conditions.

### 2.3 Step 3: The SE Knowledge Stack

In parallel with the ground-truth generation, the SE knowledge stack was developed to operationalize the high-level INCOSE criterion definitions into specific, testable rules describing how each criterion should be interpreted and applied. For each criterion, one team member captured and synthesized the reasoning the expert reviewers applied while judging requirements, supplemented it with guidance from the INCOSE and established SE references (Hull et al., 2010; INCOSE, 2023), grouped the resulting rules thematically, and abstracted them into declarative statements of the conditions a well-formed requirement should satisfy. An independent team member reviewed the resulting knowledge stack, and disagreements were resolved through discussion to reach consensus. Within MARQ, the rules are presented as a checklist beneath each criterion definition, with the evaluation logic specifying that violation of any applicable sub-rule constitutes a violation of the corresponding criterion.

Two properties of the knowledge stack are important for interpreting the results. First, the rules are generic, and they describe the conditions on requirement statements without referencing SDM, its ConOps, or any individual requirement in the evaluation set. Second, the knowledge stack and the ground truth are independent artifacts; yet they share a common origin in the expert panel's interpretation of the INCOSE criteria. The knowledge stack states that interpretation as explicit, criterion-level rules, in the same way that a codebook makes a coding scheme explicit. Whereas the ground truth is the panel's application of the same interpretation to the thirteen requirements. Our evaluation therefore measures whether an LLM equipped with an explicit expert interpretation of the criteria reproduces expert judgments made under that interpretation, which is the intended use of a knowledge-infused evaluator.

### 2.4 Step 4: MARQ Prototype Architecture

MARQ is a purpose-built, modular prototype for LLM-based requirement quality evaluation. The core idea behind MARQ is that SE industry practice and academic research will not be able to keep pace with the rapid development of commercial LLMs; hence, infusing SE knowledge in an interchangeable manner could be a productive inquiry. In other words, once formulated, a SE knowledge stack could be usable with whichever general-purpose model an organization can access, including proprietary frontier models through an API and open-weight models hosted on local infrastructure. Figure 2 shows the architecture under both experimental conditions.

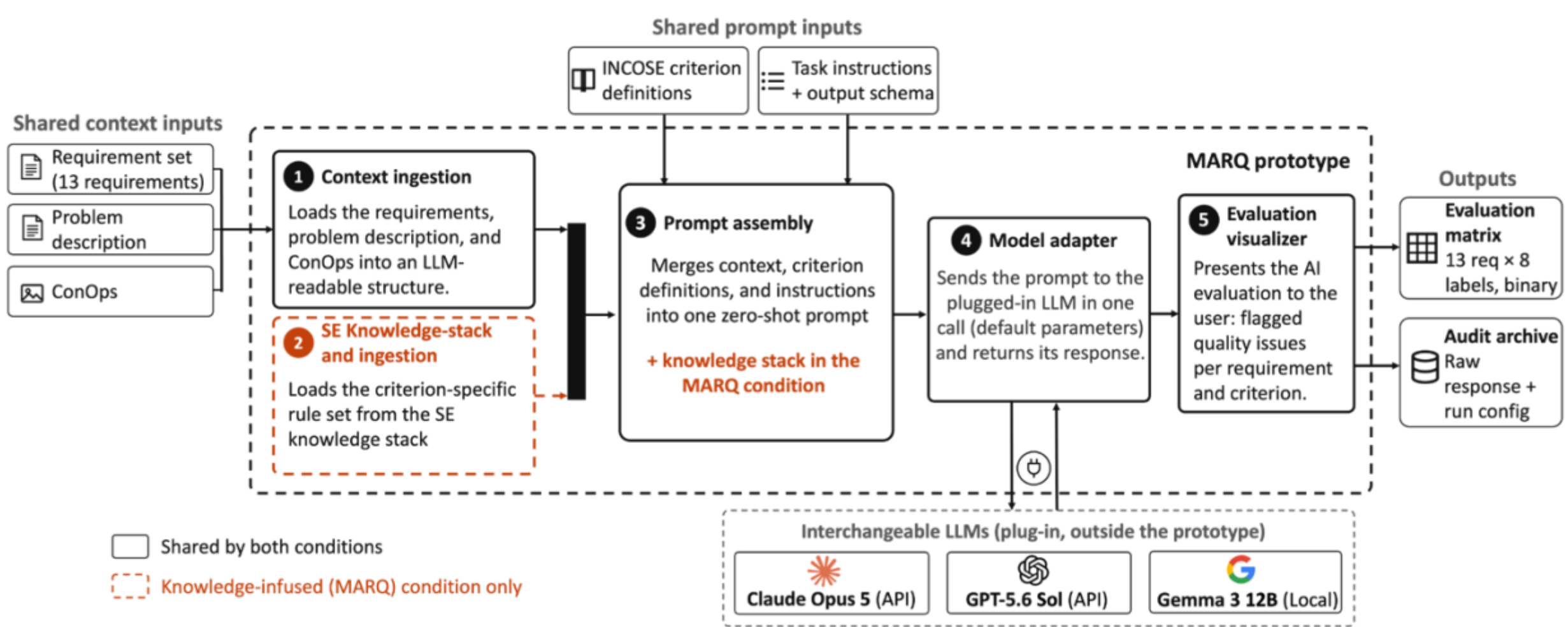

**Figure 2.** MARQ architecture under the off-the-shelf and knowledge-infused conditions.

MARQ comprises five modules: (1) *Context ingestion* loads the requirement set, problem description, and ConOps (including the ConOps figure, supplied as image content) and converts them into an LLM-readable structure. (2) *Knowledge-stack and ingestion* loads and parses the criterion-specific rule set in a parallel path. (3) *Prompt assembly* merges the shared context, the INCOSE criterion definitions, the task instructions, and the knowledge stack into a single zero-shot prompt (Kojima et al., 2022). (4) *Model adapter* submits the prompt to the selected LLM in a single inference call covering all criteria, using the model's default parameters. (5) *Evaluation visualizer* presents the AI evaluation to the user, showing both violated and non-violated criteria for each requirement. Modules 1, 2, 3, and 5 are LLM-independent, whereas switching between Claude Opus 5, GPT-5.6 Sol, and the locally hosted Gemma 3 12B requires changing only the model adapter module. The MARQ prototype is provided in the Electronic Appendix A. Next, we proceed to the experimental setup.

### 2.5 Step 5: Experimental Design and Data Collection

Two experimental conditions were defined. In the *off-the-shelf condition*, each LLM model receives the requirements, the project context and ConOps, the INCOSE criterion definitions, and the task instructions, and evaluates each requirement against each criterion. In the *MARQ condition*, the model receives exactly the same inputs plus the SE knowledge stack. The evaluation objective, output format, and model settings are identical; the sole experimental factor is the presence of the knowledge stack. The off-the-shelf condition is the reference point for both research questions.

The experiment was conducted using three LLMs: Claude Opus 5, GPT-5.6 Sol, and Gemma 3 12B. Claude Opus 5 and GPT-5.6 Sol were selected as representative frontier proprietary models available at the time of data collection. Gemma 3 12B was included as an open-weight model suitable for local deployment, with multimodal capabilities that enable it to process both textual and image-based inputs and support reasoning over contextual information relevant to requirement evaluation. Together, these models provide an intentional heterogeneous testing ground allowing the study to examine whether the effects of the prototype generalize across distinct model families and deployment paradigms rather than being specific to a single provider or model type. For each model, the off-the-shelf and prototype conditions were independently executed 100 times, resulting in six experimental configurations and 600 total evaluation runs. All LLMs were evaluated using their default model parameter settings. Repeated executions provided a means to observe both typical performance and execution-level variability to a certain extent within computational cost constraints.

No task-specific fine-tuning, retrieval augmentation, few-shot examples, or model-specific prompt optimization was used, although prior work shows such techniques can substantially affect performance (Topcu et al., 2025). This choice isolates the effect of the added knowledge stack. We note the distinction between knowledge infusion and prompt engineering explicitly. The manipulation does not change *how* the model is asked, but *what it is told about the criteria*, and the content is expert-derived, declarative, criterion-specific SE knowledge that is reusable across models and projects rather than a prompting technique tuned to a model.

Finally, both experimental conditions were implemented as automated workflows in n8n, a workflow automation platform, with separate but structurally identical workflows for the two conditions. Figure 3 provides an overview of this pipeline, showing how the pipeline maps to the MARQ architecture. Initialization starts with a time-based trigger to initiate the runs at regular intervals. Next, the workflow assigns the run configuration and execution identifier, then creates two output destinations: one for the structured requirement-by-criterion evaluation matrix and another for storing the raw model response for auditing purposes. The workflow proceeds to the next stage only after both destinations are created.

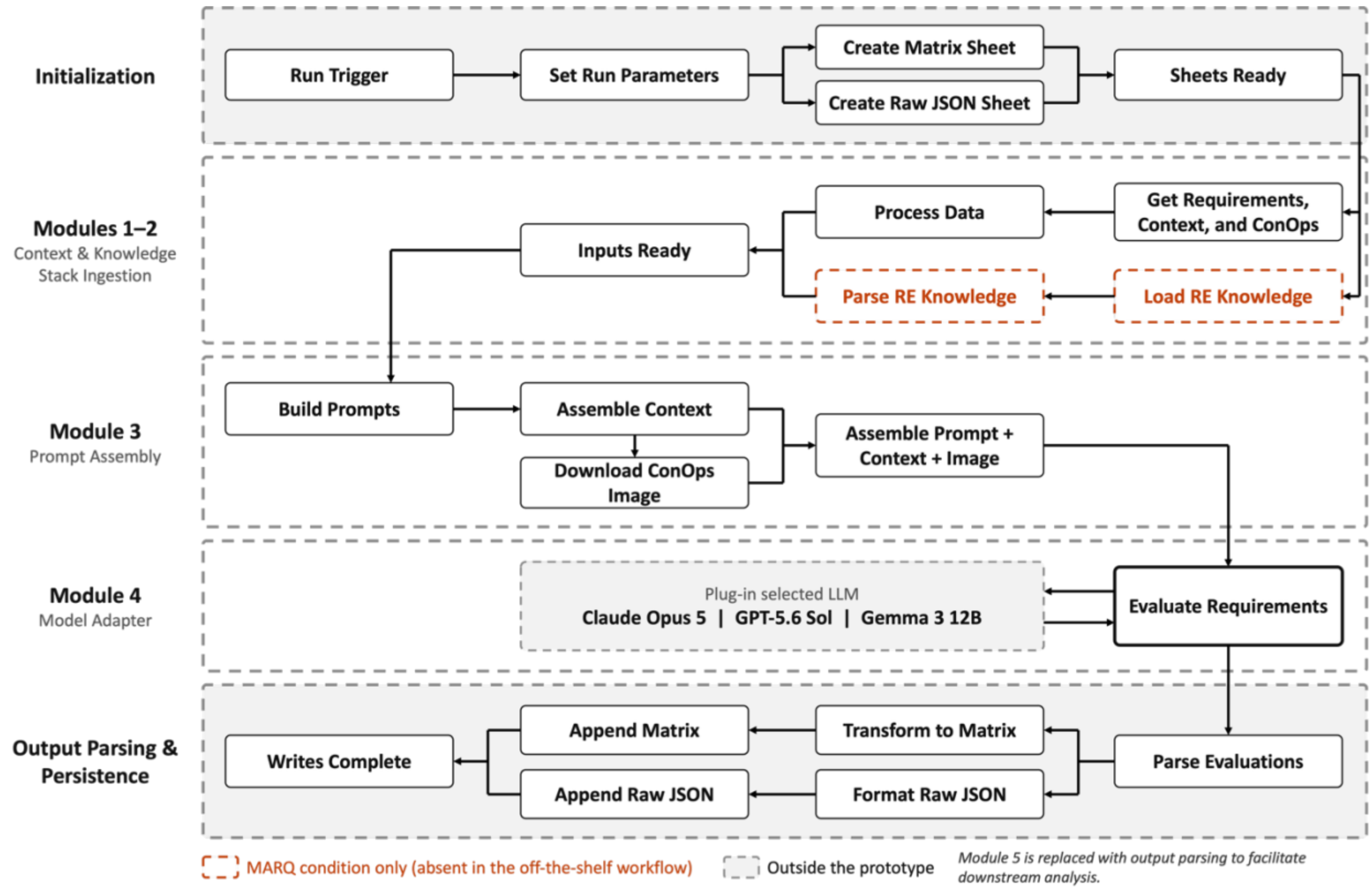

**Figure 3.** Automated LLM evaluation pipeline for Prototype Condition implemented in n8n.

During shared contextual input ingestion, the requirement set, problem description, and ConOps are loaded from source and converted into an LLM-readable structure. For the prototype condition only, a second ingestion path loads and parses the SE knowledge stack in parallel, and the two streams are synchronized before the evaluation input is constructed. The next step is to build prompts using the parsed inputs. After that, the project context and ConOps are combined into a common contextual input. The corresponding ConOps figure is retrieved and supplied directly to the model as image content. The assembled textual and visual context, as well as the prompts, is then supplied to the selected LLM model in a single inference call covering all requirement quality criteria. Finally, o*utput parsing and persistence* parses the response into a binary requirement-by-criterion matrix, stores it, and archives the raw response with the run configuration for auditing; a synchronization step confirms both are stored before the run is marked complete.

### 2.6 Step 6: Data Analysis

For each experimental run, the AI-generated judgments were first compared with the expert-derived ground truth and mapped to a standard confusion matrix (Foody, 2002; Heydarian et al., 2022), as illustrated in Figure 4:

| | | AI Evaluation | |
|---|---|---|---|
| | | **Indicates a quality issue** | **Does not indicate a quality issue** |
| **Ground Truth** | **Issue present** | True Positive (TP) | False Negative (FN) *Type II Error or Miss* |
| | **No issue** | False Positive (FP) *Type I Error or False Alarm* | True Negative (TN) . |

**Figure 4.** Confusion Matrix of AI Requirements Quality Evaluation Outcomes.

To support comparison across quality criteria and experimental conditions, performance was evaluated using normalized rates rather than raw counts. Seven metrics were calculated:

i. True positive rate (TPR) /Recall: The proportion of actual quality issues that were correctly identified by the LLM. Higher values indicate better issue-detection performance.

$$TPR = \frac{TP}{Ground\ Truth\ Total\ Positive}$$

ii. True negative rate (TNR): The proportion of true non-issues that were correctly left unflagged. Higher values indicate better non-issue classification.

$$TNR = \frac{TN}{Ground\ Truth\ Total\ Negative}$$

iii. False positive rate (FPR): The proportion of true non-issues that were incorrectly identified as quality issues. Lower values indicate better performance.

$$FPR = \frac{FP}{Ground\ Truth\ Total\ Negative}$$

iv. False negative rate (FNR): The proportion of actual quality issues that the LLM failed to detect. Lower values indicate better performance.

$$FNR = \frac{FN}{Ground\ Truth\ Total\ Positive}$$

v. Accuracy: The proportion of all classifications that were correct across both issue and non-issue cases. Higher values indicate better performance.

$$Accuracy = \frac{TP + TN}{TP + TN + FP + FN}$$

vi. Precision: The proportion of actual issues among all issues flagged by the LLM. Higher values indicate that fewer flagged cases were false alarms.

$$Precision = \frac{TP}{TP + FP}$$

vii. F1 Score: The harmonic mean of Precision and TPR, providing a single measure that balances correct issue detection with the reliability of flagged issues. A higher F1 value is better.

$$F1 = \frac{2 \times Precision \times TPR}{Precision + TPR}$$

TPR represents the LLM's ability to correctly detect existing quality issues, whereas TNR captures its ability to correctly recognize cases in which no issue is present (Luque et al., 2019). FPR and FNR are also known as Type I (false alarm) and Type II (miss), respectively (Cuellar, 2025). Accuracy summarizes the proportion of correct classifications across both classes. However, because the ground truth is class-imbalanced, accuracy alone is an incomplete summary. Therefore, precision and F1 score are considered the primary metrics of interest because they penalize false positive and false negative classifications and provide a more informative view of model performance under class imbalance. The seven performance metrics were calculated independently for every experimental run. The resulting distributions across the 100 runs were then analyzed using boxplots to examine central tendency, variability, distributional spread, and outlying observations for each LLM for each experimental condition.

## 3. Findings

This section presents the study findings in relation to the two research questions. Section 3.1 addresses RQ1 by comparing the overall performance of the off-the-shelf models and MARQ across Claude Opus 5, GPT-5.6 Sol, and Gemma 3 12B. Section 3.2 addresses RQ2 by examining criterion-level false-alarm and miss behavior. Because each experimental configuration was executed 100 times, the resulting distributions capture run-to-run variability. This variability should be treated as a characteristic of stochastic model behavior rather than experimental noise.

### 3.1 RQ1: Overall Effect of MARQ Relative to Off-the-Shelf Use

Table 2 summarizes the median and standard deviation (SD) of the seven metrics for all six configurations; Figures 5–7 show the full distributions for Claude Opus 5, GPT-5.6 Sol, and Gemma 3 12B, respectively. MARQ raises median F1 for every model, from 0.55 to 0.80 (Claude Opus 5), 0.52 to 0.62 (GPT-5.6 Sol), and 0.18 to 0.53 (Gemma 3 12B), but the nature of the gain differs, as the following subsections describe.

**Table 2.** Median (standard deviation) of performance metrics over 100 runs per configuration. Δ is the change in median from off-the-shelf (OTS) to MARQ; ↑ better / ↓ worse for the metric's direction.

| Metric | Claude Opus 5 OTS | Claude Opus 5 MARQ | Δ | GPT-5.6 Sol OTS | GPT-5.6 Sol MARQ | Δ | Gemma 3 12B OTS | Gemma 3 12B MARQ | Δ |
|---|---|---|---|---|---|---|---|---|---|
| TPR (↑) | 0.63 (0.04) | 0.83 (0.04) | +0.20 | 0.65 (0.12) | 0.90 (0.03) | +0.25 | 0.17 (0.17) | 0.57 (0.18) | +0.40 |
| TNR (↑) | 0.74 (0.04) | 0.89 (0.03) | +0.15 | 0.66 (0.04) | 0.59 (0.03) | −0.07 | 0.68 (0.09) | 0.80 (0.14) | +0.12 |
| FPR (↓) | 0.26 (0.04) | 0.11 (0.03) | −0.15 | 0.34 (0.04) | 0.41 (0.03) | +0.07 | 0.32 (0.09) | 0.20 (0.14) | −0.12 |
| FNR (↓) | 0.37 (0.04) | 0.17 (0.04) | −0.20 | 0.35 (0.12) | 0.10 (0.02) | −0.25 | 0.83 (0.17) | 0.43 (0.18) | −0.40 |
| Accuracy (↑) | 0.71 (0.03) | 0.88 (0.02) | +0.17 | 0.65 (0.04) | 0.68 (0.02) | +0.03 | 0.53 (0.04) | 0.73 (0.08) | +0.20 |
| Precision (↑) | 0.50 (0.04) | 0.76 (0.05) | +0.26 | 0.43 (0.05) | 0.47 (0.02) | +0.04 | 0.19 (0.07) | 0.53 (0.10) | +0.34 |
| F1 (↑) | 0.55 (0.03) | 0.80 (0.04) | +0.25 | 0.52 (0.07) | 0.62 (0.02) | +0.10 | 0.18 (0.10) | 0.53 (0.10) | +0.35 |

#### 3.1.1 Claude Opus 5

Figure 5 compares the performance distributions for Claude Opus 5. MARQ produces a clear improvement in detection. Median TPR rises from 0.63 (range 0.50–0.67, SD 0.04) to 0.83 (range 0.73–0.87, SD 0.04), an absolute gain of 0.20 and a relative gain of 32%. The two distributions do not overlap, so even the weakest MARQ run detects more defects than the strongest off-the-shelf run. Correspondingly, median FNR falls from 0.37 to 0.17 (−54%): the off-the-shelf model lets more than one in three genuine defects pass unflagged, whereas MARQ reduces this to roughly one in six.

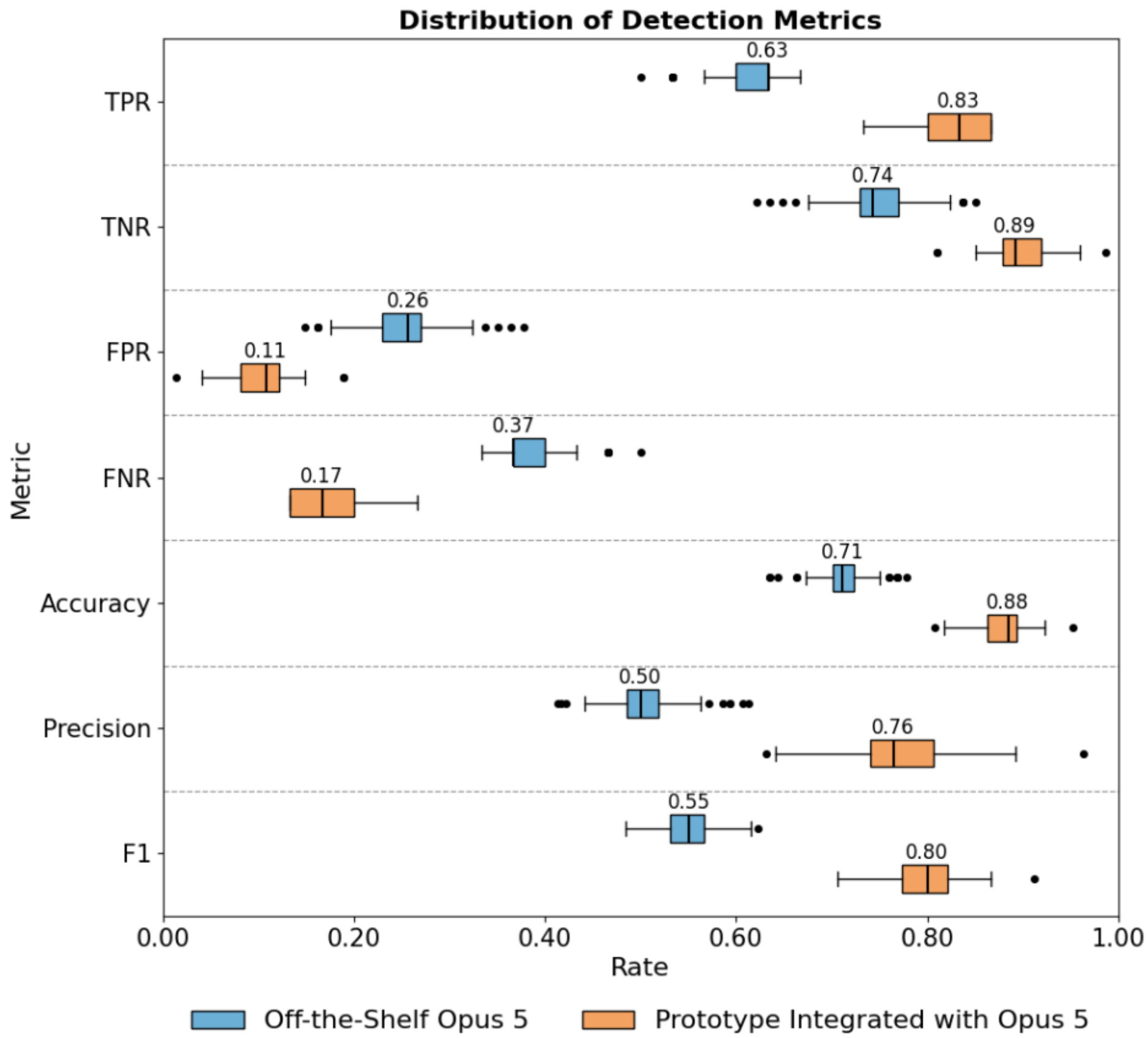

**Figure 5.** Claude Opus 5 Performance Under Off-the-Shelf and Prototype Conditions.

False-alarm behavior improves at the same time. Median FPR falls from 0.26 (range 0.15–0.38, SD 0.04) to 0.11 (range 0.01–0.19, SD 0.03), a 58% reduction, with the mirrored gain in TNR. In practical terms, the off-the-shelf model incorrectly flags roughly one quarter of sound judgments and MARQ roughly one in nine. Aggregate metrics follow as precision rises from 0.50 to 0.76 (+0.26, +52%), the largest absolute gain for this model, although its variability increases slightly (SD 0.04 to 0.05, range 0.63–0.96, with the best run approaching perfect precision). Accuracy rises from 0.71 to 0.88 (+24%) with SD decreasing from 0.03 to 0.02, and F1 rises from 0.55 to 0.80 (+45%), with SD increasing from 0.03 to 0.04 and the range widening from 0.14 to 0.21. Accuracy and F1 are fully separated across conditions. The lowest MARQ F1 (0.71) exceeds the highest off-the-shelf F1 (0.62), and the lowest MARQ accuracy (0.81) exceeds the highest off-the-shelf accuracy (0.78).

In summary, both performance in terms of true detection and avoiding false alarms improve together rather than trading against one another. Across five of the seven metrics, the worst prototype run outperforms the best off-the-shelf run; the two exceptions being the complementary FPR and TNR, which overlap marginally. In addition to that, consistency improves or remains comparable in most cases, except for precision and F1, where it declines slightly.

#### 3.1.2 GPT-5.6 Sol

Figure 6 presents the corresponding comparison for GPT-5.6 Sol. Off-the-shelf detection is unstable: median TPR is 0.65, but runs range from 0.37 to 0.83 (SD 0.12), so identical conditions can yield substantially different detection from one run to the next. MARQ raises median TPR to 0.90 (+0.25, +38%) and compresses the distribution to 0.80–0.93 (SD 0.03, a 75% reduction). Median FNR falls from 0.35 to 0.10 (−71%). Thus, MARQ significantly improves detection performance in both magnitude and variability.

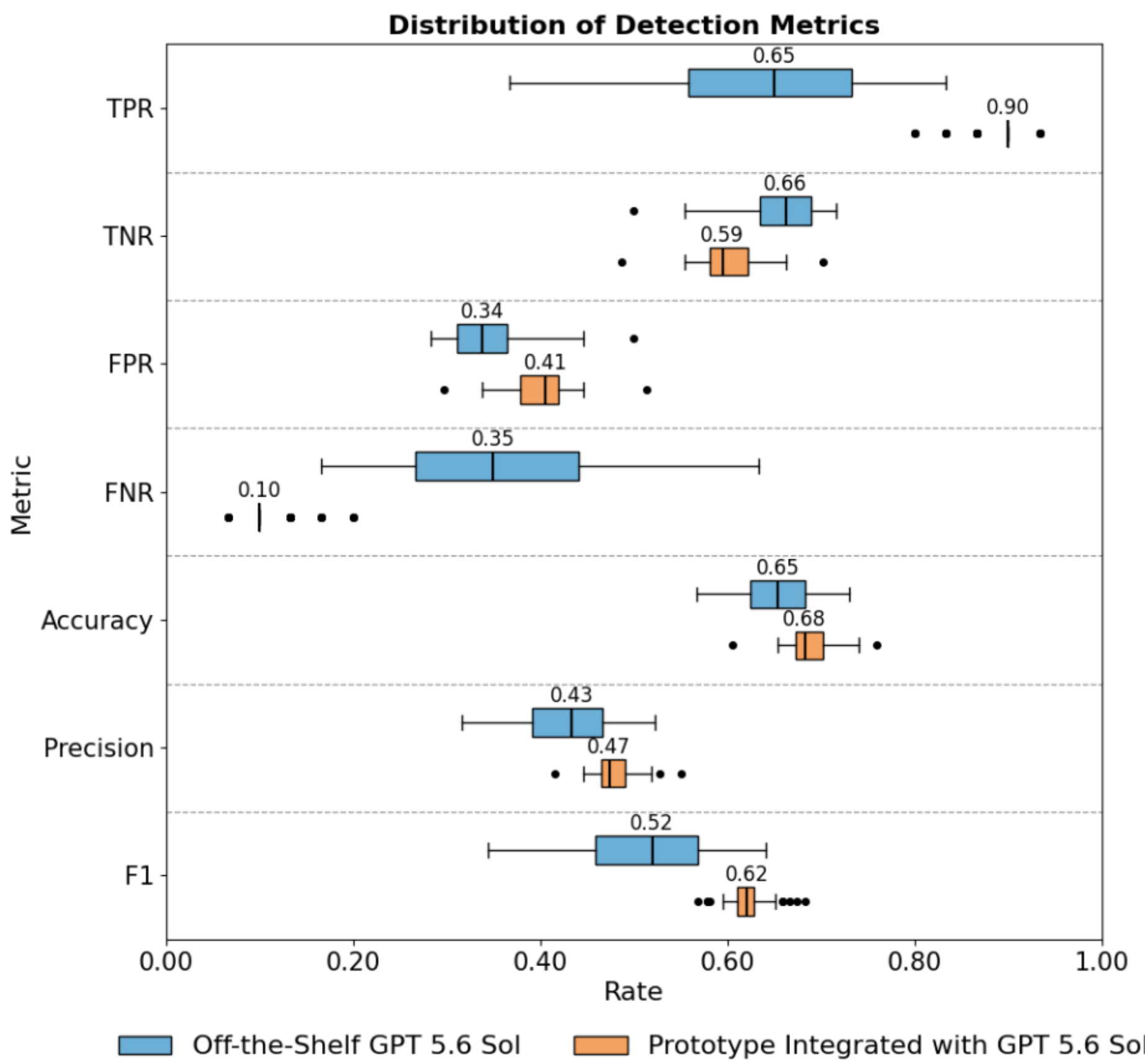

**Figure 6.** GPT-5.6 Sol Performance Under Off-the-Shelf and MARQ Conditions.

Figure 6 also reveals a trade-off. Median FPR rises from 0.34 to 0.41 (+0.07, +21%), with near-symmetric and almost fully overlapping ranges (0.28–0.50 and 0.30–0.51), and TNR falls correspondingly from 0.66 to 0.59. The off-the-shelf model already flags roughly one third of sound judgments; under MARQ this rises to roughly two in five. The strong improvement in detection is therefore accompanied by a cost in unnecessary flags. Accuracy rises only from 0.65 to 0.68 (+5%), and its variability decreases (SD 0.04 to 0.02). Median precision rises modestly from 0.43 to 0.47 (+9%); under both conditions, fewer than half of the model's flags correspond to genuine defects. F1 rises from 0.52 to 0.62 (+19%), driven almost entirely by the TPR gain.

The most pronounced effect of the prototype on GPT-5.6 Sol is therefore improved execution-level consistency rather than central tendency. Standard deviation decreases across every metric, including reductions of 75% for detection, 71% for F1, and 60% for precision. This reduction in variability is important because the usefulness of an AI depends not only on its average performance but also on whether similar results can be expected across repeated executions.

### 3.1.3 Gemma 3 12B

Figure 7 presents the performance distributions for Gemma 3 12B. Off-the-shelf detection is considerably weaker. Median TPR is only 0.17, with a wide, right-skewed distribution spanning 0.03–1.00 (SD 0.17), and median FPR (0.32) exceeds median TPR, so the model combines poor detection with a substantial false-alarm burden. MARQ raises median TPR to 0.57 (+0.40, the largest absolute gain across the three models), with the spread remaining extreme (0.20–1.00, SD 0.18); median FNR falls from 0.83 to 0.43 (−48%), although the remaining miss rate is still considerable.

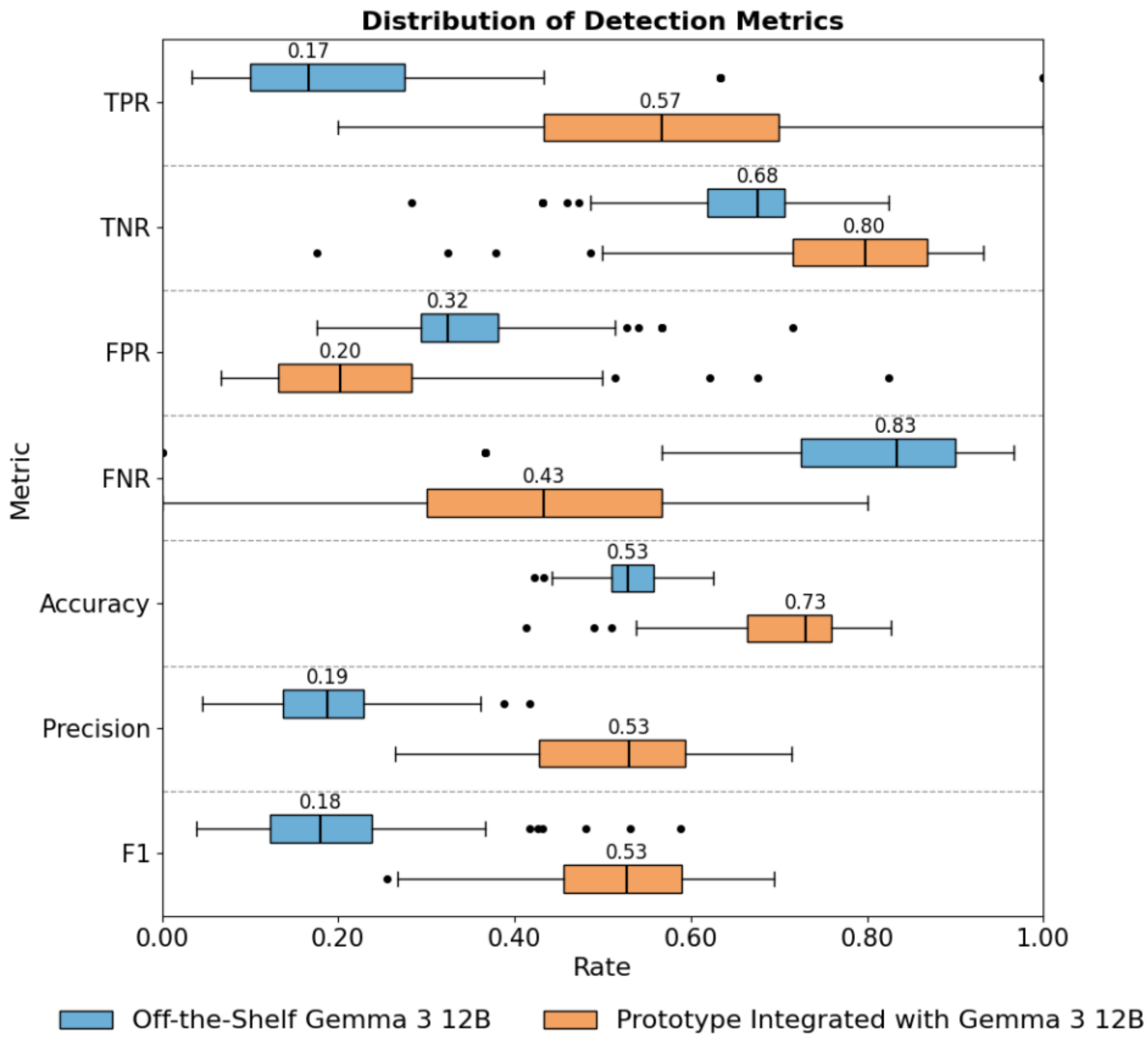

**Figure 7.** Gemma 3 12B Performance Under Off-the-Shelf and Prototype Conditions.

MARQ also improves false-alarm behavior, reducing median FPR from 0.32 to 0.20 (−38%), but FPR variability increases (SD 0.09 to 0.14, upper extreme 0.82), so some outlying MARQ runs generate excessive false alarms. Aggregate metrics show the same pattern of better typical performance with wider spread. Accuracy rises from 0.53 to 0.73 (+38%) while SD doubles (0.04 to 0.08) and the range widens from 0.42–0.63 to 0.41–0.83; precision rises from 0.19 to 0.53 (+0.34) while SD increases from 0.07 to 0.10 and the range from 0.05–0.42 to 0.26–0.71; and F1 rises from 0.18 to 0.53 (+0.35) with SD unchanged at 0.10 and the range widening from 0.04–0.59 to 0.26–0.69.

Two main observations follow. First, with MARQ, the typical performance of Gemma 3 12B becomes comparable to that of the off-the-shelf proprietary models: its median F1 of 0.53 approaches the off-the-shelf F1 of Claude Opus 5 (0.55) and GPT-5.6 Sol (0.52), and its median FPR (0.20) is lower than either (0.26 and 0.34). Second, MARQ improves central tendency for Gemma while leaving execution-level instability largely unresolved or, for most metrics, worse. Gemma remains behind the proprietary models under both conditions, and its variability remains the primary obstacle to practical use.

### 3.2 RQ2: Criterion-Level Effects of MARQ Relative to Off-the-Shelf Use

Table 3 reports median FPR and FNR by criterion, model, and condition. Figures 8–10 and Figures 11–13 show the FPR distributions and the FNR distributions, respectively. Two reading aids apply throughout. First, FNR is undefined for No Issues because the ground truth contains no positive instance, so its absence from the FNR figures carries no information about performance. Second, where a boxplot collapses to a single line at 0.00 or 1.00, the same outcome occurred in all 100 runs; an FNR collapsed at 1.00 means every issue under that criterion was missed in every run, and at 0.00 means that none was missed.

**Table 3.** Median criterion-level false positive rate (FPR) and false negative rate (FNR) over 100 runs, off-the-shelf (OTS) → MARQ. "—" = undefined.

| Criterion | Claude Opus 5 FPR | Claude Opus 5 FNR | GPT-5.6 Sol FPR | GPT-5.6 Sol FNR | Gemma 3 12B FPR | Gemma 3 12B FNR |
|---|---|---|---|---|---|---|
| Necessary | 0.17 → 0.17 | 1.00 → 0.00 | 0.25 → 0.67 | 1.00 → 0.00 | 0.83 → 0.00 | 0.00 → 1.00 |
| Appropriate | 0.00 → 0.00 | 0.50 → 0.50 | 0.09 → 0.09 | 0.50 → 0.50 | 1.00 → 0.36 | 0.00 → 1.00 |
| Unambiguous | 0.44 → 0.33 | 0.25 → 0.25 | 0.89 → 0.89 | 0.00 → 0.00 | 0.22 → 0.67 | 0.75 → 0.75 |
| Complete | 0.73 → 0.09 | 0.00 → 0.50 | 1.00 → 1.00 | 0.00 → 0.00 | 0.00 → 0.00 | 0.00 → 1.00 |
| Singular | 0.09 → 0.09 | 0.00 → 0.50 | 0.09 → 0.09 | 0.50 → 0.50 | 0.00 → 0.00 | 1.00 → 1.00 |
| Correct | 0.00 → 0.25 | 0.89 → 0.00 | 0.00 → 0.50 | 0.67 → 0.00 | 0.00 → 0.00 | 1.00 → 0.22 |
| Conforming | 0.67 → 0.00 | 0.00 → 0.10 | 0.00 → 0.00 | 0.00 → 0.10 | 0.00 → 1.00 | 1.00 → 0.00 |
| No Issues | 0.00 → 0.00 | — | 0.00 → 0.00 | — | 0.00 → 0.00 | — |

#### 3.2.1 Type I Errors or False Flags (False Positive Rates) Across Quality Criteria

Figures 8–10 compare the criterion-level FPR distributions for Claude Opus 5, GPT-5.6 Sol, and Gemma 3 12B. Overall, for some criteria, MARQ substantially reduces the tendency to incorrectly flag sound requirements; for others, false-alarm behavior remains largely unchanged or worsens.

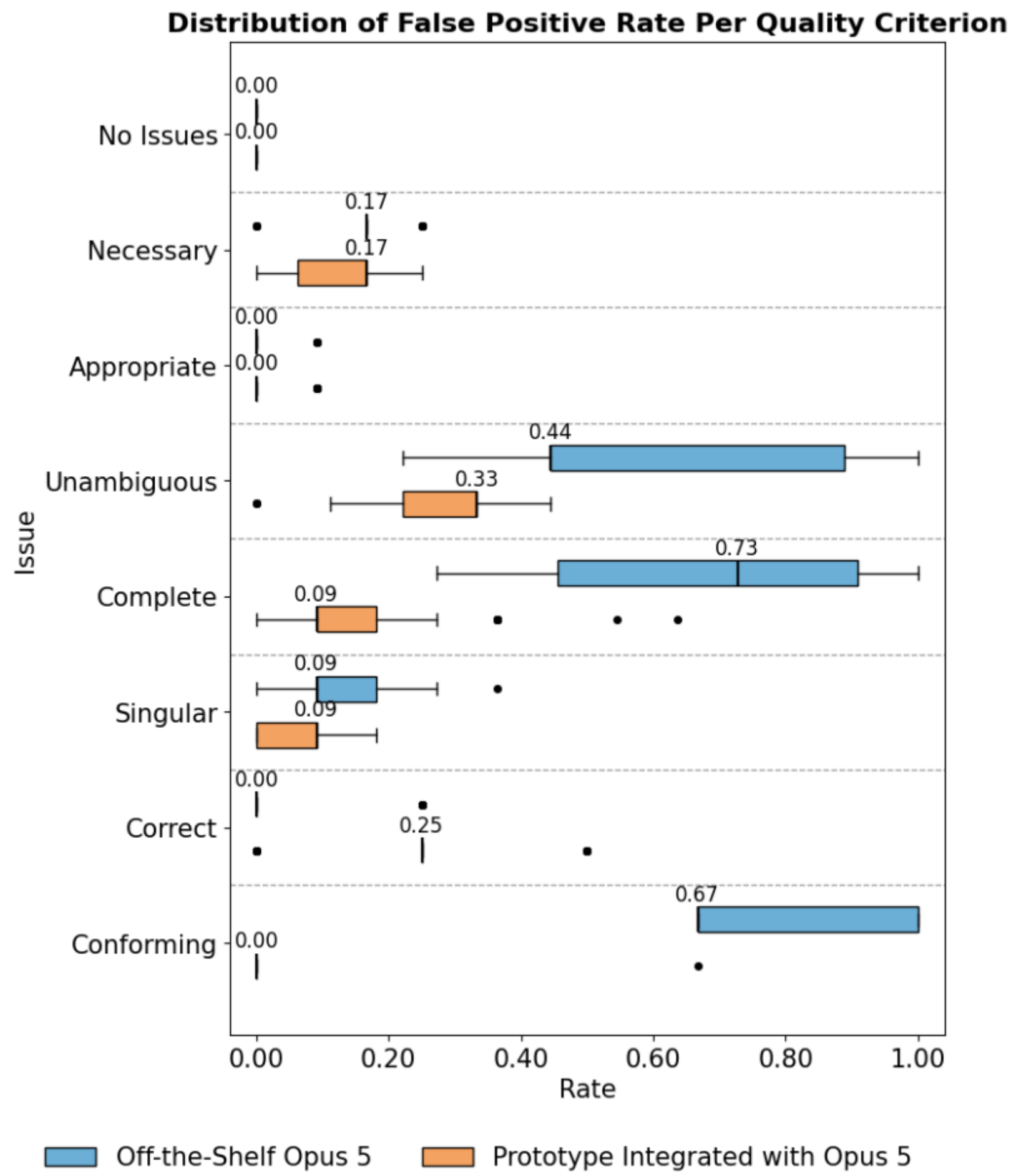

**Figure 8.** Claude Opus 5 Criterion-Level FPR Under Off-the-Shelf and Prototype Conditions.

*Claude Opus 5*. The largest reductions in false alarms occur for Conforming and Complete, the two most problematic criteria off-the-shelf. Conforming FPR falls from a median of 0.67 (mean 0.80, range 0.67–1.00, SD 0.17) to a median of 0.00 (mean 0.01); the range still extends to 0.67 because of a few outlying runs, but the distribution is extremely right-skewed. Complete FPR falls from a median of 0.73 (mean 0.67, range 0.27–1.00, SD 0.24) to 0.09 (mean 0.14, SD 0.12). Unambiguous improves moderately (median 0.44 to 0.33, mean 0.59 to 0.30, SD 0.22 to 0.11), so ambiguity judgments become both less prone to false alarms and more consistent. Singular is unchanged in median (0.09) with a lower mean (0.12 to 0.07). Necessary and Appropriate are essentially unchanged in median, although Necessary becomes more dispersed (SD 0.04 to 0.09). Correct is the exception that worsens: off-the-shelf false Correctness flags are rare (median 0.00, mean 0.02), whereas under MARQ the median rises to 0.25 (mean 0.22, range 0.00–0.50). As Section 3.2.2 shows, this increase is the price of eliminating Correctness misses.

*GPT-5.6 Sol.* A different pattern appears in Figure 9: false-alarm behavior remains high under both conditions. Conforming is the only criterion with a strong improvement; its off-the-shelf distribution has a median of 0.00 but a mean of 0.25 and a full 0.00–1.00 range (SD 0.35), whereas under MARQ FPR is 0.00 in every run. Appropriate and Singular are stable (median of 0.09 under both conditions). Complete and Unambiguous are persistent problems that MARQ does not resolve. Completeness FPR has a median of 1.00 with a range of 0.91–1.00 under both conditions, an almost invariant failure mode in which nearly every sound Completeness judgment is flagged, and Unambiguous retains a median of 0.89 (mean 0.88 to 0.83) while becoming less consistent (SD 0.05 to 0.10, range widening to 0.33–0.89). The most consequential deteriorations occur for Necessary (median 0.25 to 0.67, mean 0.28 to 0.59) and Correct, where every MARQ run produces an FPR of exactly 0.50 (two of the four sound Correctness judgments flagged) compared with an off-the-shelf median of 0.00 (mean 0.11). These two criteria explain the increase in overall FPR observed for GPT-5.6 Sol in Section 3.1.2: stronger detection is accompanied by over-flagging on criteria that require contextual judgment about whether a requirement is needed and correctly reflects the problem space.

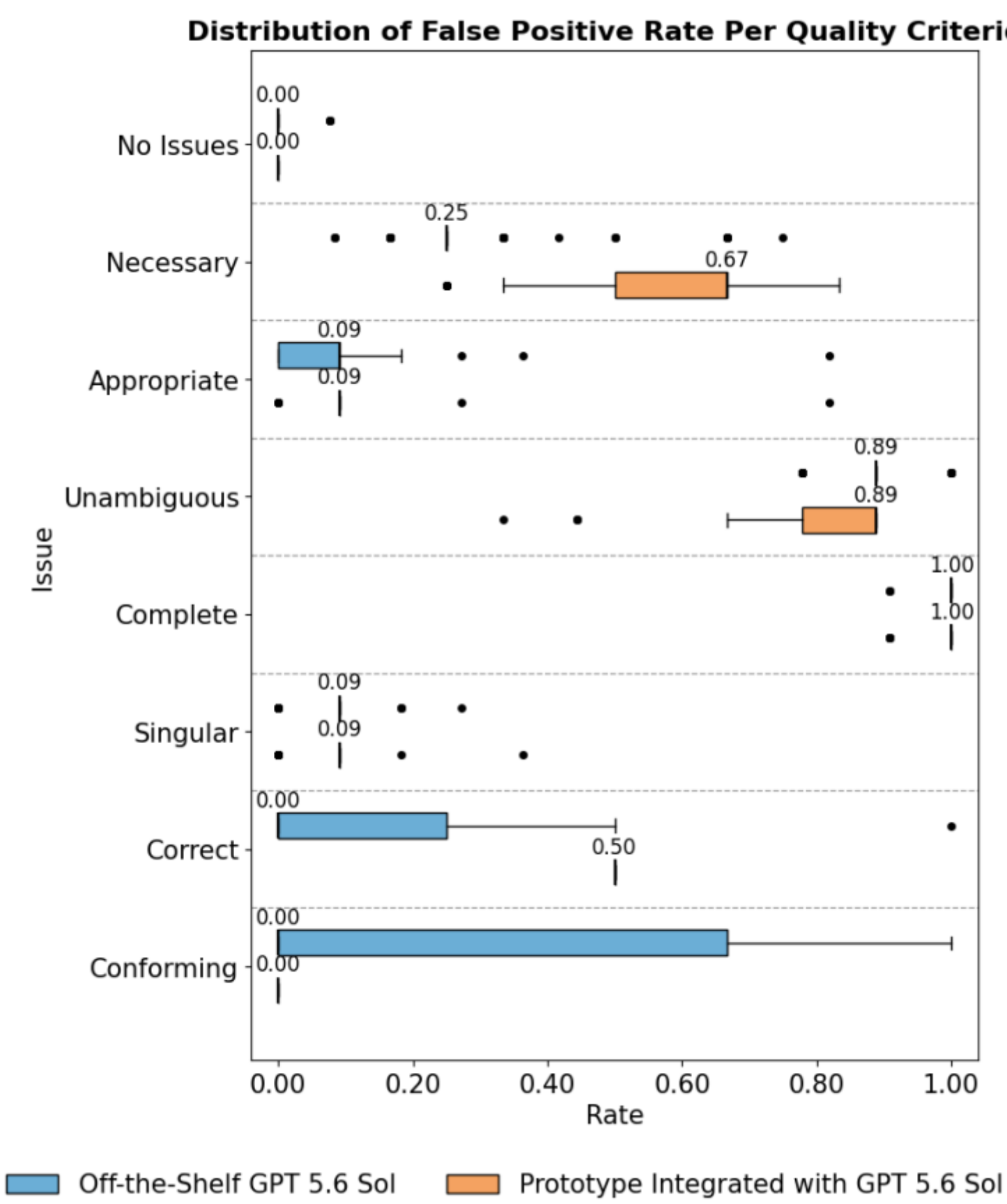

**Figure 9.** GPT-5.6 Sol Criterion-Level FPR Under Off-the-Shelf and Prototype Conditions.

*Gemma 3 12B.* Figure 10 shows a distinct third pattern. MARQ produces large false-alarm reductions for Necessary (median 0.83 to 0.00, mean 0.73 to 0.26) and Appropriate (median 1.00 to 0.36, mean 0.86 to 0.46), criteria for which the off-the-shelf model typically flags most or all sound judgments. Both improvements come with greater variability (Necessary SD 0.20 to 0.43, full 0.00–1.00 range), so the gain in typical behavior is not realized in every run. Complete and Singular remain at a median of 0.00 with lower means (Complete 0.16 to 0.01). However, MARQ worsens Unambiguous (median 0.22 to 0.67), Correct (median 0.00 under both, mean 0.02 to 0.18 with a tail to 1.00), and most strikingly Conforming, whose median moves from 0.00 to 1.00 (mean 0.22 to 0.67): in many runs every sound Conformance judgment is flagged.

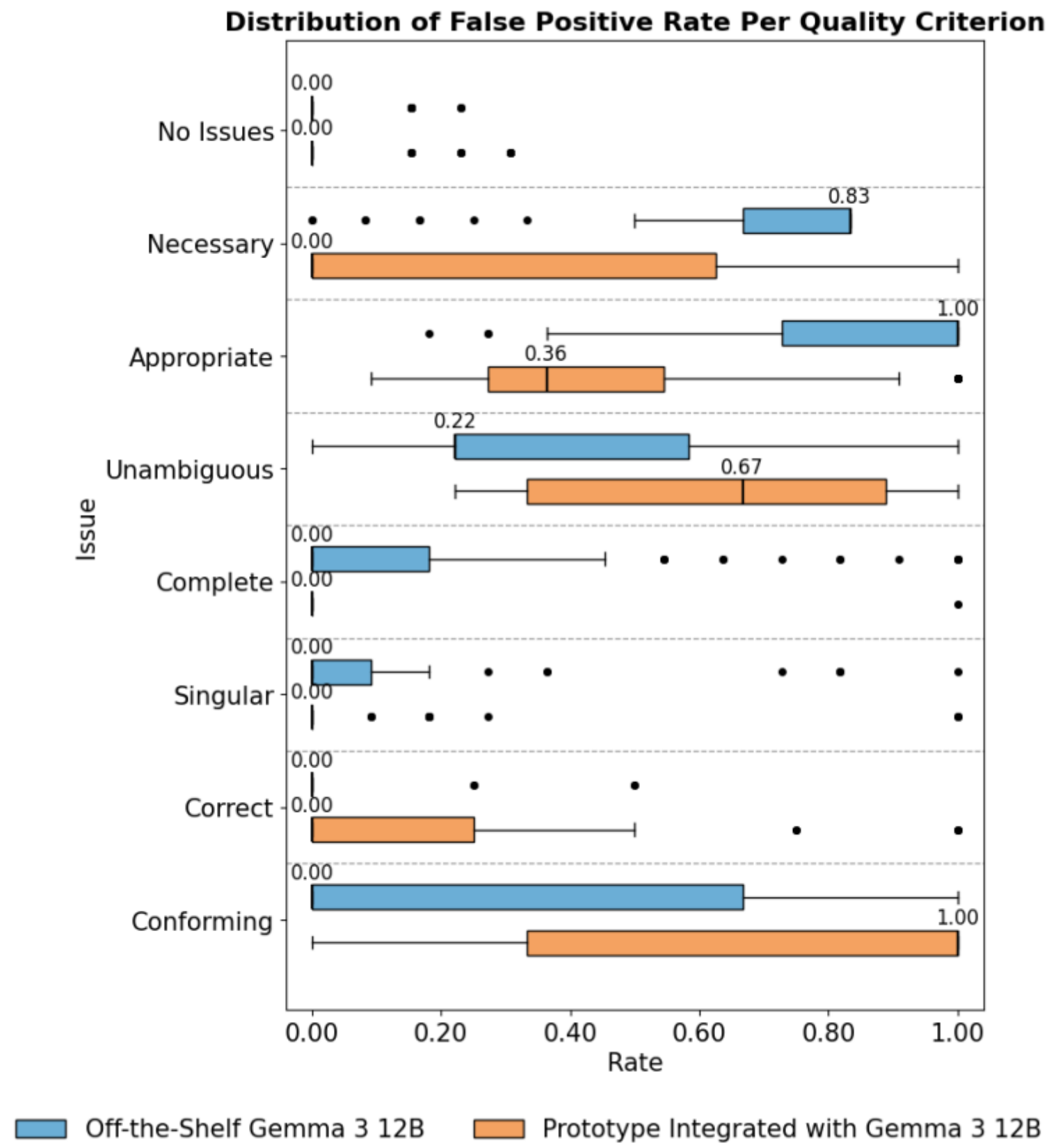

**Figure 10.** Gemma 3 12B Criterion-Level FPR Under Off-the-Shelf and Prototype Conditions.

Taken together, Figures 8–10 show that the effect of MARQ on false alarm behavior is uniform neither across INCOSE quality criteria nor across models. An overall FPR therefore needs to be read together with the criteria that generate it.

### 3.2.2 Type II Errors or Misses (False Negative Rates) Across Quality Criteria

Figures 11–13 present criterion-level FNR distributions. Similar to the false alarm behavior, the magnitude and direction of the MARQ effect on misses vary by criterion and model.

*Claude Opus 5.* The largest improvements occur for Necessary and Correct, which are severely missed off-the-shelf. The off-the-shelf model misses Necessary related issues in nearly every run (median FNR 1.00, mean 0.98), whereas MARQ detects it in all 100 runs. Correct is the more informative case: off-the-shelf FNR is 0.89 (mean 0.88, range 0.67–0.89, SD 0.03), meaning the model almost always misses most Correctness issues, whereas under MARQ, median FNR is 0.00 (mean 0.02, range 0.00–0.11). This improvement is important for SE practice because incorrect or unnecessary requirements can lead to under- or over-designed systems. Other criteria do not improve: Appropriate remains at a median FNR of 0.50 under both conditions; Unambiguous stays at a median of 0.25 with a slightly higher mean (0.18 to 0.25); and Conforming moves from a median of 0.00 (mean 0.05) to a constant 0.10 (one of ten issues missed in every run). Complete and Singular deteriorate, and MARQ moves both from a median FNR of 0.00 to 0.50, consistently for Singular (every run) and highly variably for Complete (full 0.00–1.00 range, SD 0.47). Claude Opus 5 thus shows a mixed criterion-level profile in which large gains in Necessary and Correct coexist with losses in Complete and Singular.

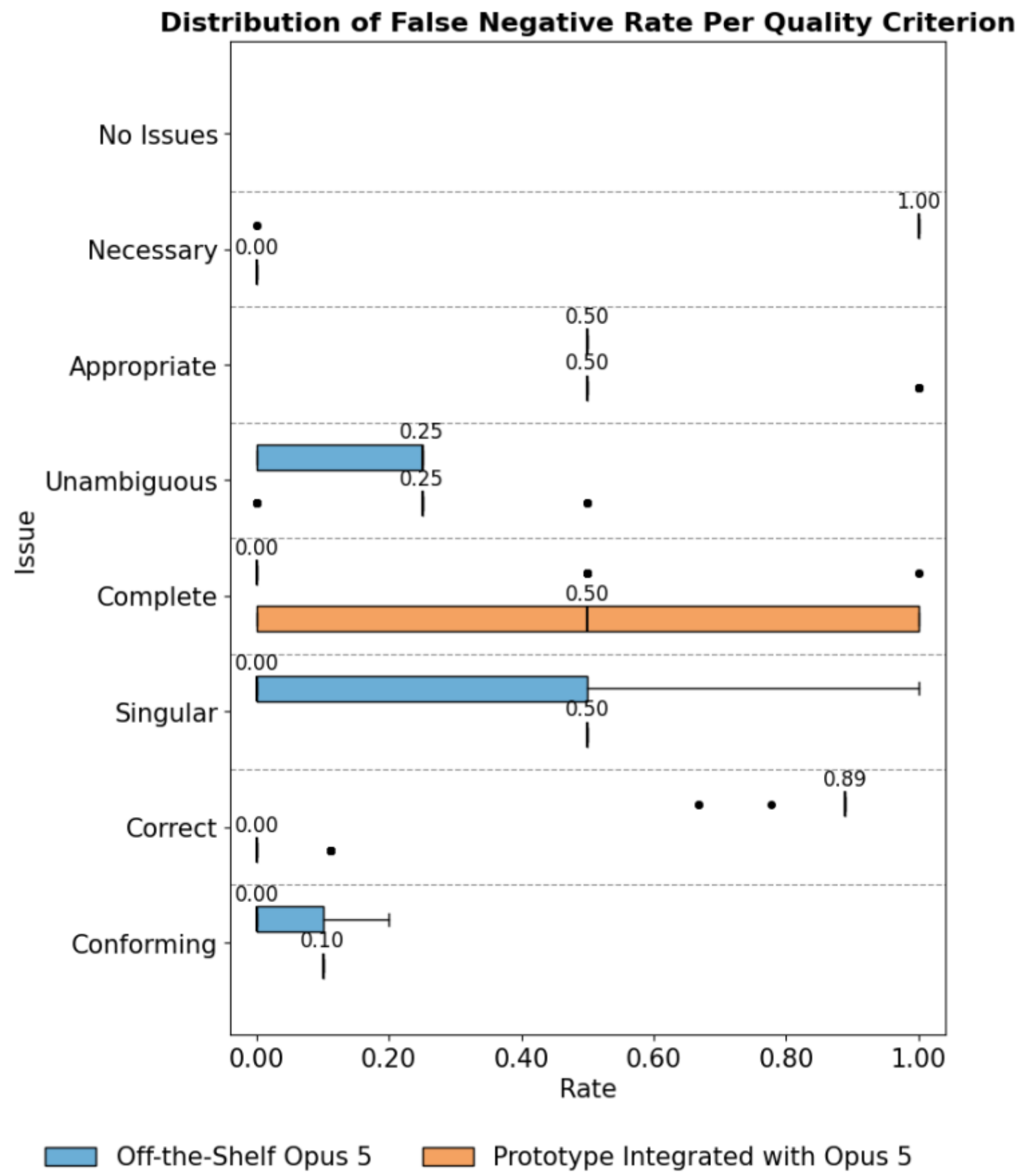

**Figure 11.** Claude Opus 5 Criterion-Level FNR Under Off-the-Shelf and Prototype Conditions.

*GPT-5.6 Sol.* As with Claude Opus 5, the strongest improvements occur for Necessary and Correct. Off-the-shelf Necessary FNR alternates between complete detection and complete failure of the single instance (median 1.00, mean 0.60, SD 0.49); under MARQ it is 0.00 in every run. Correct FNR falls from a median of 0.67 (mean 0.63, SD 0.16) to 0.00 in every run, so MARQ eliminates Correctness misses for this model. Conforming moves from a median of 0.00 (mean 0.22, with occasional failures up to 0.90) to a tightly clustered 0.10. Other criteria change little: Unambiguous remains at a median of 0.00 (mean 0.09 to 0.05); Appropriate remains at 0.50 in essentially all runs; Complete remains at a median of 0.00 (mean 0.06 to 0.08) with occasional complete misses; and Singular remains centered at 0.50 (mean 0.45 to 0.49). Unlike Claude Opus 5, no criterion deteriorates notably for GPT-5.6 Sol.

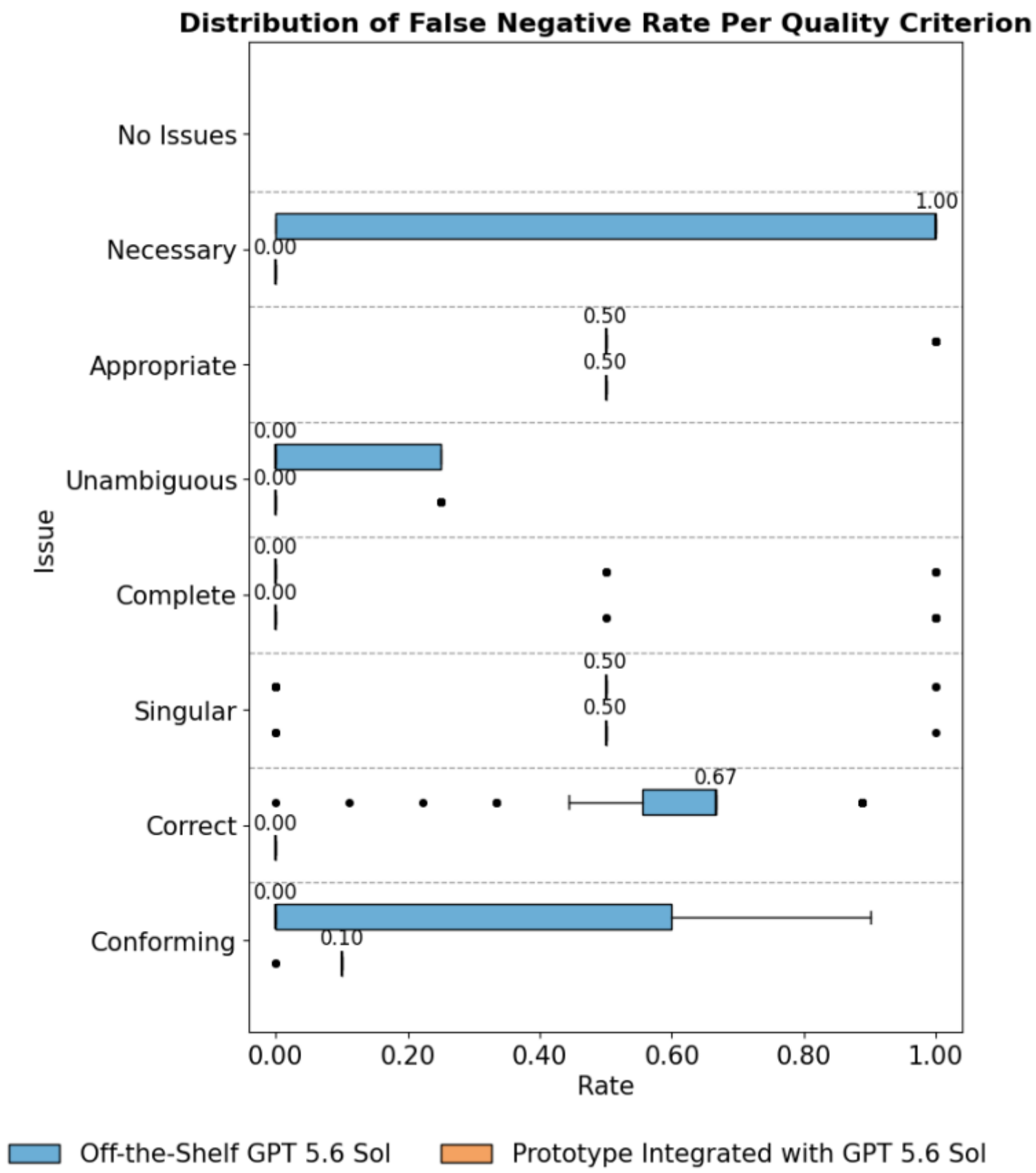

**Figure 12.** GPT-5.6 Sol Criterion-Level FNR Under Off-the-Shelf and Prototype Conditions.

*Gemma 3 12B.* Figure 13 shows the strongest Gemma improvements for Correct and Conforming. Correct FNR falls from a median of 1.00 (mean 0.97) to 0.22 (mean 0.38), although with wide variability (0.00–1.00, SD 0.33); Conforming falls from a median of 1.00 (mean 0.91) to 0.00 (mean 0.19). For these two criteria MARQ turns a model that almost always misses the defects into one that detects most of them in most runs. Unambiguous improves little (median 0.75 under both, mean 0.64 to 0.59) but becomes more consistent (SD 0.42 to 0.28). Several criteria deteriorate: Necessary moves from a median FNR of 0.00 (mean 0.26) to 1.00 (mean 0.77), so MARQ worsens detection of the single Necessary issue for Gemma while improving it for both proprietary models; Complete moves from 0.00 (mean 0.10) to 1.00 (mean 0.92); Appropriate from 0.00 (mean 0.40) to 1.00 (mean 0.53); and Singular remains at 1.00 (mean 0.95 to 0.97).

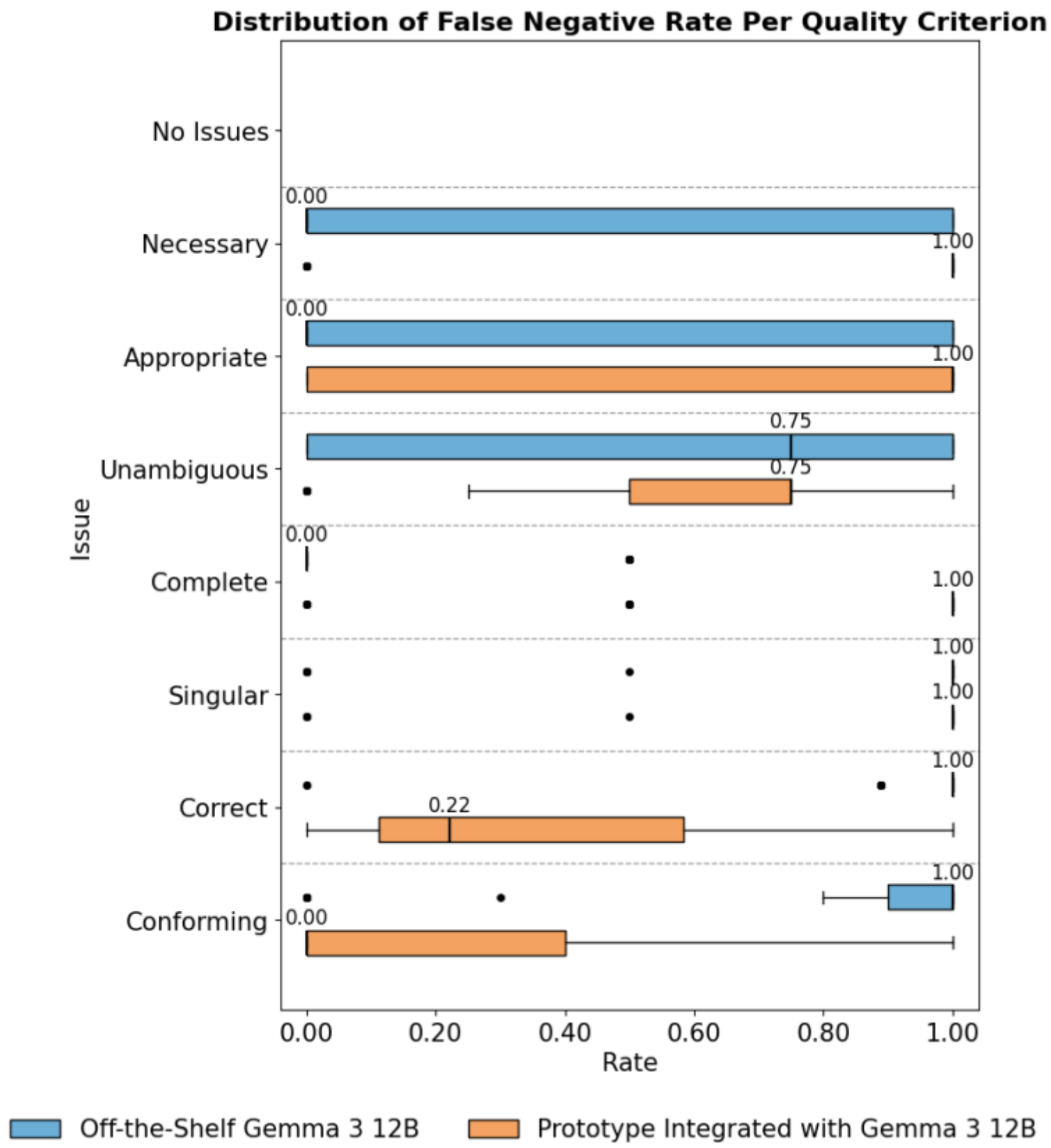

**Figure 13.** Gemma 3 12B Criterion-Level FNR Under Off-the-Shelf and Prototype Conditions.

Taken together, Figures 11–13 and Table 3 indicate that the clearest cross-model pattern is Correctness. MARQ reduces severe off-the-shelf misses for all three models, and for the two proprietary models it does so by trading a modest number of false alarms for the elimination of misses. This within-criterion trade-off is invisible in aggregate metrics and is precisely the behavior a reviewer would expect from a model that has been told what to look for.

## 4. Discussion

This study developed MARQ, a modular, LLM-agnostic prototype that infuses expert-derived SE knowledge into general-purpose LLMs, and examined its effect relative to off-the-shelf use across three models and the INCOSE quality criteria, with particular attention to whether improvement on one dimension is paid for on another. The following subsections synthesize the findings.

### 4.1 SE Knowledge Infusion Improves Every Model, but Not in the Same Way

MARQ improves overall performance for all three models relative to their own baselines, raising median F1 by 0.25 (Claude Opus 5), 0.10 (GPT-5.6 Sol), and 0.35 (Gemma 3 12B). Because the same knowledge stack produced gains for models with different architectures, providers, sizes, parameters, and baseline capability, the value of SE knowledge infusion does not appear to be an artifact of one favorable model configuration. This is the practical message of the study. SE knowledge is the transferable asset, and a modular AI prototype design could enable organizations to attach it to whichever model they can access and benefit from general advances in LLM capability without re-deriving that knowledge.

Nonetheless, the nature and magnitude of the gains are model-dependent. Claude Opus 5 shows the clearest improvement, with stronger issue detection and fewer false alarms occurring together. From a practitioner's perspective, this behavior is highly desirable because it reduces both the risk of allowing defective requirements to pass undetected and the burden created by unnecessary false flags. GPT-5.6 Sol improves substantially in detection and, above all, in execution-level consistency, with standard deviations falling by 60–75% for the primary metrics, but at the cost of more false alarms. The consistency gain is itself a trustworthiness improvement. Practitioners can expect repeated evaluations under the same conditions to yield similar outcomes. The false-alarm cost shows, however, that stronger defect detection should not automatically be read as better review support. The effect of knowledge infusion is therefore not determined by the knowledge alone; it depends on how the underlying model makes use of it.

### 4.2 A Locally Hosted Open-Weight Model Can Reach Off-the-Shelf Frontier Performance

Gemma 3 12B shows the largest relative improvement from the weakest baseline. Under MARQ, its median F1 rises from 0.18 to 0.53, comparable to the off-the-shelf F1 of Claude Opus 5 (0.55) and GPT-5.6 Sol (0.52), with a lower median false-alarm rate than either. Explicit SE knowledge thus partially compensates for weaker general-purpose capability. This finding is directly relevant to organizations that cannot send requirements to external services because of data sensitivity, export control, or infrastructure constraints, and that must therefore rely on models they can host themselves. The caveat is equally important. Gemma's execution-level variability remains extreme, and MARQ widens rather than narrows it for most metrics. Thus, while a locally hosted knowledge-infused evaluator is a credible option for typical performance, it would need additional measures, such as aggregation over repeated runs, before any single evaluation could be relied upon.

### 4.3 Gains Are Concentrated in Specific Criteria and Involve Within-Criterion Trade-offs

The criterion-level results qualify the aggregate story in three ways. First, Correctness is the clearest cross-model success: all three models miss most Correctness issues off-the-shelf, and MARQ substantially reduces those misses for all three. Correctness is one of the criteria the benchmark identified as a priority target because it depends on linking a requirement to the problem statement and ConOps rather than on local textual cues (Shefa et al., 2026). The result suggests that stating explicitly what such a judgment requires helps LLMs make context-dependent judgments they otherwise fail to make. Second, that help comes with a within-criterion trade-off for the proprietary models: Correctness false alarms rise from near zero to one-quarter and one-half of sound judgments. A knowledge stack that tells a model what to look for makes it look harder, and the appropriate balance between the two error types is a deployment decision, not a property of the tool. Third, because Correct and Conforming account for nearly two-thirds of the ground-truth positives, the aggregate gains largely reflect these two criteria.

These differences nonetheless reveal where the next capability development effort should go. A single intervention does not affect every criterion in the same way, and criteria such as Singular and Conforming, whose violations are largely structural, may be better served by complementary rule-based or ontology-supported methods (Chen et al., 2020; Dermeval et al., 2016) than by further knowledge infusion, whereas criteria requiring contextual judgment appear to be where knowledge infusion earns its keep. A modular architecture makes such criterion-specific routing feasible.

### 4.4 Implications for Trustworthy AI for SE

The findings reinforce that an LLM-based evaluator should be judged not by a single aggregate metric but by its misses, false alarms, criterion-level behavior, and execution-level consistency, because improvements on one dimension can mask deterioration on another. They also show that even the strongest model–MARQ combination does not approach perfect performance on any metric, and that identically configured evaluations can still produce different outcomes. The most defensible role for MARQ is therefore as an initial reviewer that targets expert attention. MARQ can surface likely problems, provide a structured second opinion, and make review more efficient, while human reviewers validate flags, resolve ambiguous cases, and interpret requirement quality in the broader system context. The same conclusion extends to emerging agentic workflows in which a requirement quality evaluator is one component of a larger SE pipeline. More capable and consistent components improve downstream decisions, but model- and criterion-dependent behavior means each component should be validated for its specific role rather than assumed to be uniformly reliable.

## 5. Conclusion, Limitations, and Future Work

This study asked whether infusing expert-derived SE knowledge into general-purpose LLMs improves requirement quality assessment beyond off-the-shelf capability, and whether that improvement transfers across model families. It presented MARQ, a modular, LLM-agnostic prototype built around a reusable, criterion-specific knowledge stack, and evaluated it against off-the-shelf baselines using two frontier proprietary models and one open-weight model over a total of 600 runs. MARQ improved median F1 for every model, improved detection of Correctness issues for every model, made one proprietary model markedly more consistent, and brought a locally hostable 12B-parameter model to the typical performance of off-the-shelf frontier models. At the same time, the kind of improvement depended on the model; gains were concentrated in Correctness and Conformance, and several criteria showed within-criterion trade-offs or model-dependent deterioration.

The broader implication is that infusing SE knowledge is a promising and model-agnostic way of turning general advances in LLMs into dependable RE support. However, it is not a complete solution, and knowledge-infused evaluators belong in human-in-the-loop architectures and workflows. Capabilities such as MARQ can therefore be used to prioritize reviewer attention, provide a structured initial or second opinion, and support more efficient quality assurance, while final engineering judgment remains with the human expert reviewer. These implications extend beyond standalone requirement-review tools to emerging agentic AI workflows, where requirement-quality evaluators may influence subsequent SE activities and therefore should be validated for their intended function.

Several limitations should be acknowledged. The evaluation is based on a relatively small engineered system, so the findings may not fully generalize to larger or more organizationally complex systems, different domains, or requirement sets with different maturity levels. As more complex systems typically impose greater contextual and systems-level reasoning demands, present results should be interpreted as a comparatively favorable scenario. Also, requirement quality evaluation is treated as a binary classification task, which does not capture dimensions such as defect severity, explanation quality, or partially correct judgments. In addition, the tested models do not represent the full range of available commercial, local, or restricted LLMs. Taken together, the reported performance values should not be interpreted as universally representative of all AI-assisted requirement quality evaluation settings, although the broader patterns observed are expected to remain relevant across engineered system contexts. It is worth noting that the study does not incorporate optimization strategies such as prompt engineering, retrieval augmentation, fine-tuning, domain-specific instruction tuning, or structured reasoning. This was an intentional design choice to isolate the effect of the expert knowledge infusion, and attribute observed performance differences as directly as possible to the added knowledge rather than to other LLM performance augmentation techniques.

Future research should test MARQ on larger and more diverse requirement datasets, additional engineered-system domains, and broader model families. Further work should also investigate richer and criterion-specific knowledge representations, including ontologies, knowledge graphs, and retrieval-augmented approaches, and exploit MARQ's modularity to route different criteria to different evaluation strategies. Controls for prompt length and structure, and aggregation schemes over repeated runs to tame execution-level variability, are natural methodological extensions. Human-AI studies are also needed to determine whether the substantial model-level improvements observed in this study translate into more efficient, reliable, and trustworthy requirement review in practice.

**Acknowledgments:** This research was supported by the National Nuclear Security Administration (NNSA) through Systems Engineering Research Center (SERC) Contract WRT-2416: Systems Engineering Beyond the Horizon. Any views, opinions, findings, conclusions, or recommendations expressed in this material are those of the authors and do not necessarily reflect the views of NNSA.

**Data Availability Statement:** The data that support the findings of this study are available from the corresponding author upon reasonable request.

**Conflict of interest:** The authors declare no conflicts of interest.

## References

Arora, C., Grundy, J., & Abdelrazek, M. (2024). Advancing requirements engineering through generative ai: Assessing the role of llms. In *Generative AI for Effective Software Development* (pp. 129–148). Springer.

Ashish, V. (2017). Attention is All you Need. *Advances in Neural Information Processing Systems*, *30*, I.

Aysan, O., Akyol, G., & Tüzün, E. (2026). *Detecting Software Requirements Smells Through Domain-Augmented Generation*. 334–345.

Baker, J. (1975). The DRAGON system–An overview. *IEEE Transactions on Acoustics, Speech, and Signal Processing*, *23*(1), 24–29.

Boehm, B., Valerdi, R., & Honour, E. (2008). The ROI of systems engineering: Some quantitative results for software-intensive systems. *Systems Engineering*, *11*(3), Article 3. https://doi.org/10.1002/sys.20096

Boehm, B. W., & Papaccio, P. N. (1988). Understanding and controlling software costs. *IEEE Transactions on Software Engineering*, *14*(10), Article 10.

Chen, R., Chen, C.-H., Liu, Y., & Ye, X. (2020). Ontology-based requirement verification for complex systems. *Advanced Engineering Informatics*, *46*, 101148.

Cheng, H., Husen, J. H., Lu, Y., Racharak, T., Yoshioka, N., Ubayashi, N., & Washizaki, H. (2026). Generative ai for requirements engineering: A systematic literature review. *Software: Practice and Experience*, *56*(2), 141–170.

Courbis, A.-L., Wei, J., & Lambolais, T. (2026). ELISA, a requirements ELIcitation Smart Agent—Comparison of state-of-the-art LLMs. *ACM Transactions on Software Engineering and Methodology*.

Cuellar, M. (2025). The problem with eliminations: Why forensic comparisons need false negative rates. *Forensic Science International: Synergy*, *11*, 100621. https://doi.org/10.1016/j.fsisyn.2025.100621

Cui, J., & Alexander, R. (2026). Same Prompt, Different Outcomes: Evaluating the Reproducibility of Data Analysis by LLMs. *arXiv Preprint arXiv:2602.14349*.

Dalpiaz, F., Ferrari, A., Franch, X., & Palomares, C. (2018). Natural language processing for requirements engineering: The best is yet to come. *IEEE Software*, *35*(5), 115–119.

Dermeval, D., Vilela, J., Bittencourt, I. I., Castro, J., Isotani, S., Brito, P., & Silva, A. (2016). Applications of ontologies in requirements engineering: A systematic review of the literature. *Requirements Engineering*, *21*(4), 405–437.

Elman, J. L. (1990). Finding Structure in Time. *Cognitive Science*, *14*(2), 179–211. https://doi.org/10.1207/s15516709cog1402_1

Fantechi, A., Gnesi, S., Passaro, L., & Semini, L. (2023). *Inconsistency detection in natural language requirements using chatgpt: A preliminary evaluation*. 335–340.

Fazelnia, M., Koscinski, V., Herzog, S., & Mirakhorli, M. (2024). *Lessons from the use of natural language inference (nli) in requirements engineering tasks*. 103–115.

Foody, G. M. (2002). Status of land cover classification accuracy assessment. *Remote Sensing of Environment*, *80*(1), 185–201.

Hayes-Roth, F. (1985). Rule-based systems. *Communications of the ACM*, *28*(9), 921–932. https://doi.org/10.1145/4284.4286

Hazelrigg, G. (1998). A Framework for Decision-Based Engineering Design. *Journal of Mechanical Design*, *120*(4), Article 4. https://doi.org/10.1115/1.2829328

Heesch, R., Eilermann, S., Windmann, A., Diedrich, A., & Niggemann, O. (2025). *Evaluating large language models for real-world engineering tasks*. 54–66.

Heydarian, M., Doyle, T. E., & Samavi, R. (2022). MLCM: Multi-label confusion matrix. *Ieee Access*, *10*, 19083–19095.

Hochreiter, S., & Schmidhuber, J. (1997). Long short-term memory. *Neural Computation*, *9*(8), 1735–1780.

Hu, M., Zhou, Y., Fan, W., Nie, Y., Xia, B., Sun, T., Ye, Z., Jin, Z., Li, Y., & Chen, Q. (2025). Owl: Optimized workforce learning for general multi-agent assistance in real-world task automation. *arXiv Preprint arXiv:2505.23885*.

Hull, E., Jackson, K., & Dick, J. (2010). Requirements engineering in the solution domain. In *Requirements Engineering* (pp. 115–136). Springer.

INCOSE. (2023). *INCOSE systems engineering handbook*. John Wiley & Sons. https://books.google.com/books?hl=en&lr=&id=HhjBEAAAQBAJ&oi=fnd&pg=PR9&dq=Guide+to+Writing+Requirements+incose&ots=hPDzRao1nR&sig=mG4PIX63Hcafz02u3TM6dGhe7EU

Izhar, R., Bhatti, S. N., & Alharthi, S. A. (2025). Bridging precision and complexity: A novel machine learning approach for ambiguity detection in software requirements. *IEEE Access*, *13*, 12014–12031.

Jafari, S. M., Yildirim, S., Cevik, M., & Basar, A. (2023). *Anaphoric ambiguity resolution in software requirement texts*. 4722–4730.

Jelinek, F., Mercer, R. L., Bahl, L. R., & Baker, J. K. (1977). Perplexity—A measure of the difficulty of speech recognition tasks. *The Journal of the Acoustical Society of America*, *62*(S1), S63–S63.

Jin, D., Sun, W., Huang, J., Liang, P., Xuan, J., Liu, Y., & Jin, Z. (2025). iredev: A knowledge-driven multi-agent framework for intelligent requirements development. *arXiv Preprint arXiv:2507.13081*.

Kaddour, J., Harris, J., Mozes, M., Bradley, H., Raileanu, R., & McHardy, R. (2023). Challenges and applications of large language models. *arXiv Preprint arXiv:2307.10169*.

Kassab, M., Neill, C., & Laplante, P. (2014). State of practice in requirements engineering: Contemporary data. *Innovations in Systems and Software Engineering*, *10*(4), 235–241.

Kojima, T., Gu, S. S., Reid, M., Matsuo, Y., & Iwasawa, Y. (2022). Large language models are zero-shot reasoners. *Advances in Neural Information Processing Systems*, *35*, 22199–22213.

Lubos, S., Felfernig, A., Tran, T. N. T., Garber, D., El Mansi, M., Erdeniz, S. P., & Le, V.-M. (2024). *Leveraging llms for the quality assurance of software requirements*. 389–397.

Luitel, D., Hassani, S., & Sabetzadeh, M. (2023). Using Language Models for Enhancing the Completeness of Natural-Language Requirements. In A. Ferrari & B. Penzenstadler (Eds.), *Requirements Engineering: Foundation for Software Quality* (Vol. 13975, pp. 87–104). Springer Nature Switzerland. https://doi.org/10.1007/978-3-031-29786-1_7

Luque, A., Carrasco, A., Martín, A., & de Las Heras, A. (2019). The impact of class imbalance in classification performance metrics based on the binary confusion matrix. *Pattern Recognition*, *91*, 216–231.

Mahaju, S., Carver, J. C., & Bradshaw, G. L. (2023). Human error management in requirements engineering: Should we fix the people, the processes, or the environment? *Information and Software Technology*, *160*, 107223.

Mahbub, T., Dghaym, D., Shankarnarayanan, A., Syed, T., Shapsough, S., & Zualkernan, I. (2024). Can GPT-4 aid in detecting ambiguities, inconsistencies, and incompleteness in requirements analysis? A comprehensive case study. *IEEE Access*, *12*, 171972–171992.

Malik, G., Cevik, M., Başar, A., & Parikh, D. (2025). Supervised semantic similarity-based conflict detection algorithm: S3CDA. *2025 IEEE International Conference on Collaborative Advances in Software and COmputiNg (CASCON)*, 1–10.

Melo, W. (2026). Improving the system requirement engineering process with artificial intelligence. *2026 IEEE International Systems Conference (SysCon)*, 1–3.

Norheim, J. J., Rebentisch, E., Xiao, D., Draeger, L., Kerbrat, A., & Weck, O. L. de. (2024). Challenges in applying large language models to requirements engineering tasks. *Design Science*, *10*, e16. https://doi.org/10.1017/dsj.2024.8

Nuseibeh, B., & Easterbrook, S. (2000). *Requirements engineering: A roadmap*. 35–46.

Peña, M., & Valerdi, R. (2015). Characterizing the Impact of Requirements Volatility on Systems Engineering Effort. *Systems Engineering*, *18*(1), 59–70. https://doi.org/10.1111/sys.21288

Poulsen, V. V., Guertler, M., Eisenbart, B., Tomidei, L., & Sick, N. (2026). Context-aware large language models for ambiguity detection in requirements. *Proceedings of the Design Society*, *6*, 2493–2502.

Salado, A., & Nilchiani, R. (2014). A Categorization Model of Requirements Based on Max-Neef's Model of Human Needs. *Systems Engineering*, *17*(3), 348–360. https://doi.org/10.1002/sys.21274

Sapkota, R., Roumeliotis, K. I., & Karkee, M. (2025). Ai agents vs. Agentic ai: A conceptual taxonomy, applications and challenges. *Information Fusion*, 103599.

Shefa, J., Salado, A., Wach, P., & Topcu, T. G. (2026). Two Truths and A Lie? Benchmarking Off-the-Shelf LLMs for Requirements Quality Assessment: Performance, False Alarms, and Misses. *arXiv Preprint arXiv:2609.03230*.

Sikdar, A., & Payyazhi, J. (2014). A process model of managing organizational change during business process redesign. *Business Process Management Journal*, *20*(6), 971–998.

Simon, H. A. (1973). The structure of ill structured problems. *Artificial Intelligence*, *4*(3–4), 181–201.

Simon, H. A. (1995). Problem forming, problem finding and problem solving in design. *Design & Systems*, 245–257.

Singh, A., & Szajnfarber, Z. (2025). Architecting Human-AI Systems for Effective Collaboration and Oversight: Making Sense of Human/AI-in/on/Over/Under/Along-the-Loop. *Systems Engineering*, *n/a*(n/a), e70024. https://doi.org/10.1002/sys.70024

Sonbol, R., Rebdawi, G., & Ghneim, N. (2022). The use of nlp-based text representation techniques to support requirement engineering tasks: A systematic mapping review. *Ieee Access*, *10*, 62811–62830.

Stecklein, J. M., Dabney, J., Dick, B., Haskins, B., Lovell, R., & Moroney, G. (2004, June 19). *Error Cost Escalation Through the Project Life Cycle*. 14th Annual International Symposium, Toulouse. https://ntrs.nasa.gov/citations/20100036670

Sutskever, I., Vinyals, O., & Le, Q. V. (2014). Sequence to sequence learning with neural networks. *Advances in Neural Information Processing Systems*, *27*. https://proceedings.neurips.cc/paper/2014/hash/a14ac55a4f27472c5d894ec1c3c743d2-Abstract.html

Szajnfarber, Z., Hennig, A., Mukherjee, S., Rader, S., & Crusan, J. (2023). Linking solver characteristics, solving processes and solution attributes: A data explainer for an open innovation generated robotic design dataset. *Data in Brief*, *50*, 109547.

Szajnfarber, Z., Zhang, L., Mukherjee, S., Crusan, J., Hennig, A., & Vrolijk, A. (2020). Who is in the crowd? Characterizing the capabilities of prize competition competitors. *IEEE Transactions on Engineering Management*, *69*(4), 1537–1551.

Talha, F., Tahir, T., & Nadeem, T. (2025). A Semiautomated Approach for Detecting Ambiguities in Software Requirements Using SpanBERT and Named Entity Recognition. *Journal of Software: Evolution and Process*, *37*(8), e70041.

Topcu, T. G., Husain, M., Ofsa, M., & Wach, P. (2025). Trust at Your Own Peril: A Mixed Methods Exploration of the Ability of Large Language Models to Generate Expert-Like Systems Engineering Artifacts and a Characterization of Failure Modes. *Systems Engineering*, sys.21810. https://doi.org/10.1002/sys.21810

Van Lamsweerde, A. (2000). *Requirements engineering in the year 00: A research perspective*. 5–19.

Vasudevan, P., & Reddivari, S. (2025). *The role of generative ai models in requirements engineering: A systematic literature review*. 188–194.

Vincenti, W. G. (1990). *What engineers know and how they know it* (Vol. 141). Baltimore: Johns Hopkins University Press.

Zhang, S., Yin, M., Zhang, J., Liu, J., Han, Z., Zhang, J., Li, B., Wang, C., Wang, H., & Chen, Y. (2025). Which agent causes task failures and when? On automated failure attribution of llm multi-agent systems. *arXiv Preprint arXiv:2505.00212*.

Zhao, L., Alhoshan, W., Ferrari, A., Letsholo, K. J., Ajagbe, M. A., Chioasca, E.-V., & Batista-Navarro, R. T. (2022). Natural Language Processing for Requirements Engineering: A Systematic Mapping Study. *ACM Computing Surveys*, *54*(3), 1–41. https://doi.org/10.1145/3444689

Zhou, X., Wang, X., He, Y., Zou, R., Wu, Y., Cheng, Y., Xie, Y., Liu, W., Zhao, H., & Xu, Y. (2026). *Engibench: A benchmark for evaluating large language models on engineering problem solving*. 36308–36334.

**Electronic Appendix A: MARQ Prototype Repository**

To support transparency, reproducibility, and further research in the field, the MARQ prototype and supporting documentation are available through the MARQ GitHub Repository.